\documentclass[final,3p,euclid]{elsarticle}

\usepackage{amssymb}
\RequirePackage[l2tabu,orthodox]{nag}
\usepackage{microtype}
\usepackage{amssymb}
\usepackage{graphicx}
\usepackage{subfigure}
\usepackage{amsmath}
\allowdisplaybreaks[4]
\usepackage{amsfonts}
\usepackage{amsbsy}
\usepackage{bm}
\usepackage{array}
\usepackage{times}
\usepackage{color}
\usepackage{ulem}
\usepackage{lineno}
\usepackage{multirow}
\usepackage{mathrsfs}
\usepackage{booktabs}
\usepackage[colorlinks,linkcolor=red,anchorcolor=blue,citecolor=green]{hyperref}
\usepackage{cleveref}
\usepackage{booktabs}
\usepackage{siunitx}
\usepackage{fmtcount}
\usepackage[noend]{algpseudocode}
\usepackage{algorithmicx,algorithm}

\usepackage{pifont}
\usepackage{threeparttable}
\usepackage{tikz}
\usetikzlibrary{shapes.geometric,backgrounds,fit,arrows,positioning}
\usepackage{titlesec}
\titleformat{\paragraph}
{\itshape}{\theparagraph.}{0.7em}{}
\titlespacing*{\paragraph}
{0pt}{3.25ex plus 1ex minus .2ex}{0ex}
\usepackage{indentfirst}
\usepackage{lineno}
\usepackage{xcolor}
\newcolumntype{R}{>{\color{red}}l}
\begin{document}
	\begin{frontmatter}
		\noindent \textcolor{red}{This manuscript has been published in\textit{\textcolor{red}{Reliability Engineering \& System Safety}} with DOI:}  \href{https://doi.org/10.1016/j.ress.2024.110309}{\textcolor{red}{10.1016/j.ress.2024.110309}}. \\
		\textcolor{red}{ Please cite as: C. Ding, P. Wei, Y. Shi, J Liu, M. Broggi, M. Beer, Sampling and active learning methods for network reliability estimation using K-terminal spanning tree, Reliability Engineering \& System Safety (2024) 110309. }

		
		
		\title{Sampling and active learning methods for network reliability estimation using K-terminal spanning tree}
		
		
		\author[label1]{Chen Ding}
		\address[label1]{ Institute for Risk and Reliability, Leibniz University Hannover, Callinstr. 34, Hannover 30167, Germany}

		\author[label2]{Pengfei Wei\corref{cor1}}
		\ead{pengfeiwei@nwpu.edu.cn}
		\address[label2]{School of Power and Energy, Northwestern Polytechnical University, Dongxiang Road 1, Xi’an 710129, Shaanxi Province, P.R. China}
		
		\author[label1]{Yan Shi}
		
		\author[label1]{Jinxing Liu}
		
		\author[label1]{Matteo Broggi}
		
		\author[label1,label3,label4]{Michael Beer}
	
		\address[label3]{Institute for Risk and Uncertainty, University of Liverpool, Peach Street, Liverpool L69 7ZF, United Kingdom}
		\address[label4]{International Joint Research Center for Resilient Infrastructure \& International Joint Research Center for Engineering Reliability and Stochastic Mechanics, Tongji University, Shanghai 200092, PR China}
	

		
		
		\cortext[cor1]{Corresponding author}

\begin{abstract}
Network reliability analysis remains a challenge due to the increasing size and complexity of networks. 
This paper \textcolor{black}{presents a novel sampling method and an active learning method} for efficient and accurate network reliability estimation under node failure and edge failure scenarios. 
The proposed \textcolor{black}{sampling} method adopts Monte Carlo technique to sample component lifetimes and the K-terminal spanning tree algorithm to accelerate structure function computation. 
\textcolor{black}{Unlike existing methods that compute only one structure function value per sample, our method generates multiple component state vectors and corresponding structure function values from each sample.
Network reliability is estimated based on survival signatures derived from these values. 
A transformation technique extends this method to handle both node failure and edge failure.
To enhance efficiency of proposed sampling method and achieve adaptability to network topology changes, we introduce an active learning method utilizing a random forest (RF) classifier.}
This classifier directly predicts structure function values, integrates network behaviors across diverse topologies, and undergoes iterative refinement to enhance predictive accuracy. 
Importantly, the trained RF classifier can directly predict reliability for variant networks, a capability \textcolor{black}{beyond the sampling method alone. }
Through investigating several network examples \textcolor{black}{and two practical applications}, the effectiveness of both proposed methods is demonstrated.

\end{abstract}


\end{frontmatter}


\section{Introduction}\label{}
In modern society, infrastructure networks with complex topologies, such as electricity distribution, railroad, mobile and communication networks, are ubiquitous.  
The safe and reliable operation of these networks is essential for the well-being of society. 
Any disruption, whether from natural disasters like earthquakes or from aging, poses significant risks, leading to economic losses and threats to community safety \cite{dehghani2021adaptive}. 
Therefore, reliability analysis of infrastructure networks is required to determine the probability that a network with unreliable components will work as expected under specified functional conditions \cite{paredes2019principled}.
Despite extensive research in this area, the escalating size and complexity of networks lead to an increase in the computational costs associated with such analyses. 
Consequently, efficient network reliability analysis remains an opening challenge. 

The classical methods for computing network reliability can be broadly divided into two categories: methods for calculating exact reliability and methods for estimating reliability. 
The most direct exact method entails the enumeration of states and the enumeration of paths or cuts for computation, but the computational complexity of this method grows exponentially with network complexity \cite{lucet1999exact}. 
In this regard, many improved exact methods are developed to enhance the computational efficiency.
For example, the sum of disjoint method transforms the structure function into a sum of disjoint products using inversion techniques, simplifying the evaluation of network reliability based on minimal path or cut sets \cite{chaturvedi2002efficient,cacscaval2017sdp}.
The state-space decomposition method decomposes the system state space into multiple state subsets and obtains network reliability through the probability of the subset union \cite{doulliez1972transportation,bai2018reliability}.
The binary decision diagram (BDD) method constructs a BDD by encoding and manipulating the Boolean functions to compactly represent the network reliability function, and evaluates network reliability by traversing the BDD structure \cite{hardy2007k}.
However, due to the NP-hard nature of exact network reliability computation, these exact methods are challenging to apply to large-scale and general-structured networks.
As an alternative, approximation methods have gained traction for estimating network reliability in large-scale networks. 
Elperin et al. \cite{elperin1991estimation}, for instance, integrates Monte Carlo simulation (MCS) with maximum spanning tree analysis, utilizing graph evolution models to compute network reliability.
Stern et al. \cite{stern2017accelerated} employs MCS to generate component failure samples, and constructs a support vector machine and logistic regression based surrogate model to substitute the exact node connectivity check within the MCS method for approximating the network reliability.
Lee et al. \cite{lee2023efficient} utilizes the subset simulation in estimating two-terminal reliability and obtaining the network fragility curve of lifeline networks, where two piecewise continuous functions are proposed to reformulate the binary network limit state function.

In recent years, novel approaches to network reliability analysis have emerged, one of which involves leveraging the concept of survival signature \cite{coolen2012generalizing}.
Survival signature is an extension from the system signature to accommodate multiple classes of non-identically distributed component lifetimes.
The key advantage of survival signature lies on its ability to separate the network topology structure from the probability information of component lifetimes, facilitating a lossless compression of the system structure function.
Once the survival signature is computed, network reliability analysis becomes efficient, enabling consideration of various scenarios such as imprecise uncertainty propagation \cite{feng2016imprecise,patelli2017simulation,salomon2021efficient}, dependencies modeling \cite{george2019extending,bai2021statistical} and common causes of failure \cite{mi2020reliability}.
However, the computation of survival signature is susceptible to the curse of dimensionality, leading to rapidly increasing computational requirements with the growth of the number of components and network complexity.
To address this challenge, several methods have been developed.
For example, Reed et al. \cite{reed2019efficient} combines the BDD, boundary set partition sets and simple array operations to calculate exact survival signatures in the case of only edges failing.
Nonetheless, constructing BDDs requires a significant amount of memory as the number of components and component types increases.
Another approach proposed by Behrensdorf et al. \cite{behrensdorf2021numerically} integrates percolation theory and Monte Carlo simulation (MCS) to approximate survival signatures for undirected networks with unreliable nodes. 
While this method reduces memory requirements by avoiding the evaluation of structure functions for all survival signature entries, its accuracy depends on the number of MCS samples, limiting its applicability to larger and more complex networks.
Furthermore, Di Maio et al. \cite{di2023entropy} develops an entropy-based MCS method to improve computational efficiency by using entropy to guide sampling in unknown survival signature regions. 
However, this method may face challenges when applied to networks with a larger number of unreliable nodes.

To extend applicability to \textcolor{black}{larger} networks with enhanced accuracy and efficiency, surrogate model-based and optimization-based methods can be employed to approximate survival signature and assess network reliability.
For instance, Di Maio et al. \cite{di2024ensemble} treats survival signature approximation as a missing data problem and constructs an ensemble of artificial neural networks (ANNs) to predict missing values of survival signature based on sparse data given by MCS.
Similarly, Behrensdorf et al. \cite{behrensdorf2023imprecise} forms a training dataset by collecting key values of survival signature using MCS, builds a normalized radial basis function (NRBF) network upon the training dataset to approximate the survival signature and extends the NRBF network to an interval predictor model to account for approximation errors.
Additionally, Lopes da Silva and Sullivan \cite{da2023optimization} discover the relationship between the survival signature computation and multi-objective optimization, and propose a bi-objective optimization-based MCS method to compute survival signatures in two-terminal networks with two-class unreliable nodes.
While these methods have demonstrated efficacy in approximating survival signatures for large-scale real-world networks with two component classes, they face limitations in direct application to variant networks, such as those with deleted components, due to neglecting the intricate relationships and interactions between the network behavior and different network topology structures.
Moreover, these methods often focus solely on node failure scenarios in their investigations, neglecting the applicability to edge failure scenarios. 
\textcolor{black}{Besides, these methods adopt network connectivity algorithms, such as the shortest path algorithm, which can evaluate only one structure function value per component state vector (network topology). 
Recognizing the above shortcomings, there arises a need for versatile and comprehensive approaches that accommodate different variant networks and effectively handle both node failure and edge failure scenarios.
Moreover, an efficient network connectivity algorithm capable of computing multiple structure function values from a single network topology analysis is essential for further advancement.}

\textcolor{black}{In this paper, we introduce two novel methods for network reliability analysis, particularly focusing on evaluating two-terminal reliability for networks with binary-state components. 
While our primary focus is on two-terminal networks, the proposed methods are versatile and can be adapted to K-terminal networks as well. 
The contribution of proposed methods is to enhance efficiency in network connectivity checks, address both node failure and edge failure scenarios effectively and adapt seamlessly to different variant networks. 
First, we propose a novel MCS method based on the K-terminal spanning tree (MC-KST). This method employs the Monte Carlo technique to sample component lifetimes and adopts a new scheme named K-terminal spanning tree algorithm to efficiently compute structure function values.
Here, we assume lifetimes of each component of the same class are independently and identically distributed.
Unlike existing MCS-based methods, the MC-KST obtains a set of component state vectors from each lifetime sample, and a set of structure function values associated with each sample can be determined by running the K-terminal spanning tree algorithm once. 
Network reliability is then assessed based on the survival signature approximated from the obtained component state vectors and structure function values.
Additionally, the MC-KST includes a transformation technique enabling the seamless adaptation from node failure scenarios to edge failure scenarios.
However, directly applying the MC-KST to variant networks poses challenges. 
To address this, we introduce an active learning method based on K-terminal spanning tree (AL-KST), an extension of the MC-KST that employs a new active learning strategy.
AL-KST adopts a random forest (RF) classifier trained on a small number of MCS samples to predict structure function values, facilitating seamless integration of diverse network behaviors and topology structures. 
A learning function is developed to effectively drive the iterative refinement process of the RF classifier, enhancing its prediction accuracy.
Once properly trained, the RF classifier can be directly applied to variant networks without additional computational burden, thus improving adaptability and efficiency in network reliability analysis.}

\textcolor{black}{The remaining of the paper is organized as follows.
Section 2 provides an overview of network reliability estimation. 
Sections 3 and 4 introduce the proposed MC-KST and AL-KST methods, respectively, for evaluating two-terminal reliability in network systems with binary-state components. 
Section 5 presents six synthetic network examples and two real-world network cases, illustrating the feasibility of the proposed methods under both node failure and edge failure scenarios. 
Finally, Section 6 offers discussion and concluding remarks. }

\section{Problem statement} \label{section:section2}
Consider a network described by an undirected \textcolor{black}{connected} graph $\mathcal{G} =\left( \mathcal{V} , \mathcal{E} , \mathcal{T} \right)$, where $\mathcal{V} =\left( v_1,...,v_{n_v} \right) $ denotes the set of $n_v$ nodes, $\mathcal{E} = \left({e_1, ..., e_{n_e}}\right)$ is the set of $n_e$ edges, and $\mathcal{T} \subseteq \mathcal{V}$ is a set of $\mathcal{K}$ special nodes that serve as terminals.
\textcolor{black}{In this network, let $\mathcal{M}$ represent the number of components prone to failure. 
These components are assumed to be in a binary state, and their states are described by a Boolean vector $\boldsymbol{X}=\left( X_1,X_2,...,X_{\mathcal{M}} \right) \in \left\{ 0,1 \right\} ^\mathcal{M} $, where $X_i=1$ indicates the $i$-th component is operational while $X_i=0$ signifies a failed component state.
The failure of each component is determined by a specified probability function that describes its lifetime (or failure time), which is assumed to be independently distributed for each component.
We assume all terminal nodes in $\mathcal{T}$ are perfectly reliable, and their connectivity determines the network state such that the network functions if there exists a path through working nodes and edges among all terminal nodes, and the network fails if no such path exists \cite{reed2019efficient}. 
To describe the network state, a structure function $\phi \left( \boldsymbol{X} \right) \in \left\{ 0,1 \right\} $ is defined, with $\phi \left( \boldsymbol{X} \right)=1$ denoting a functional network and $\phi \left( \boldsymbol{X} \right)=0$ indicating a failed network.
}

\textcolor{black}{Suppose the network include $\mathcal{S}$ classes of components, where components within the same class share an identical lifetime distribution.
The number of components of the $s$-th class is denoted by $\mathcal{M}_s$, and we have $\sum_{s=1}^{\mathcal{S}}{\mathcal{M} _s=}\mathcal{M}$.
The component state vector can then be redefined as $\boldsymbol{X}=\left( \boldsymbol{X}^1,\boldsymbol{X}^2,...,\boldsymbol{X}^{\mathcal{S}} \right) $, with $\boldsymbol{X}^s=\left( X_{1}^{s},X_{2}^{s},...,X_{\mathcal{M} _s}^{s} \right), s=1,2,...,\mathcal{S}$ representing the state vector of the class-$s$ components. 
Let $C_{\mathcal{M} _s}^{l_s}$ represent the total number of component state vectors with precisely $l_s$ of $\mathcal{M}_s$ class-$s$ components functioning, where $\sum_{i=1}^{\mathcal{M} _s}{X_{i}^{s}=l_s},  s=1,2,...,\mathcal{S}$. 
The set containing all component state vectors in the network is denoted as $\mathcal{H}_{l_1,l_2,...,l_{\mathcal{S}}}$ with size $\prod_{s=1}^{\mathcal{S}}{C_{\mathcal{M} _s}^{l_s}}$. 
The survival signature, denoted by $\Phi \left( l_1,l_2,...,l_{\mathcal{S}} \right)$, refers to the probability that network functions under the condition that precisely $l_s$ of its class-$s$ components are operational \cite{coolen2012generalizing}}:
\begin{equation}\label{eq:SurvivalSignature}
	\Phi \left( l_1,l_2,...,l_{\mathcal{S}} \right) ={{\sum_{\boldsymbol{X}\in \mathcal{H}_{l_1,l_2,...,l_{\mathcal{S}}}}{\phi \left( \boldsymbol{X} \right)}}\Bigg/{\left( \prod_{s=1}^{\mathcal{S}}{C_{\mathcal{M} _s}^{l_s}} \right)}}.
\end{equation}

Suppose there are $L_s\left( t \right) \in \left\{ 0,1,...,\mathcal{M} _s \right\}$ class-$s$ components functioning at time \textcolor{black}{$t > 0 $}, and the \textcolor{black}{common} cumulative distribution function (CDF) of the lifetimes of class-$s$ components, denoted as $F_s \left( t \right)$, is known.
Then the probability that exactly $l_s \in \left\{0,1,...,\mathcal{M}_s \right\} $ components of each component class function at time $t$ is computed by
\begin{equation}\label{eq:ComponentProbability}
	P\left( \bigcap_{s=1}^{\mathcal{S}}{\left\{ L_s\left( t \right) =l_s \right\}} \right) =\prod_{s=1}^{\mathcal{S}}{P\left( L_s\left( t \right) =l_s \right)}=\prod_{s=1}^{\mathcal{S}}{C_{\mathcal{M} _s}^{l_s}\left[ F_s\left( t \right) \right] ^{\mathcal{M} _s-l_s}\left[ 1-F_s\left( t \right) \right] ^{l_s}}.
\end{equation} 
The network reliability (or network survival function) describing the probability that the network functions at time $t$ can be defined according to the law of total probability, i.e.,
\begin{equation}\label{eq:NetworkReliability}
	R\left( t \right) =P\left\{ T_F>t \right\} =\sum_{l_{1}=0}^{\mathcal{M} _1}{\cdots \sum_{l_{\mathcal{S}} =0 }^{\mathcal{M} _{\mathcal{S}}}{\Phi \left( l_1,l_2,...,l_{\mathcal{S}} \right) P\left( \bigcap_{s=1}^{\mathcal{S}}{\left\{ L_s\left( t \right) =l_s \right\}} \right)}},
\end{equation}
where $T_F$ is the lifetime of network.
Note from Eq. (\ref{eq:SurvivalSignature}) that the survival signature only depends on the network topology structure and is not related to the lifetimes of components. 
Besides, once the survival signature is obtained, the network reliability can be computed by Eq. (\ref{eq:NetworkReliability}) without additional computational efforts.
 
In this paper, we focus on estimating two-terminal network reliability, which is the probability \textcolor{black}{that there exists at least one operational path connecting two terminal nodes (i.e., the source node and the target node).}
Two types of component failure are considered in this study: (1) edge failure, i.e., only edges in set $\mathcal{E}$ may fail and all nodes in set $\mathcal{V}$ are completely reliable, \textcolor{black}{with $\mathcal{M}=n_e$}; (2) node failure, i.e., only nodes in set $\mathcal{V}$ may fail and all edges in set $\mathcal{E}$ are fully reliable, \textcolor{black}{with $\mathcal{M}=n_v$}. 
Note that the analytical solution of survival signature depends on obtaining the structure function values for all $2^\mathcal{M}$ possible component state vectors.
\textcolor{black}{The structure function values can be computed by using existing connectivity check algorithms (such as the Dijkstra algorithm \cite{dijkstra2022note}) to determine the existence of path from source node to target node. }
Nevertheless, enumerating all the component state vectors and calculating the corresponding structure function can be extremely time-demanding, especially for large-scale complex networks.
Although some \textcolor{black}{sampling methods, such as MCS based methods \cite{behrensdorf2021numerically},} can be applied to approximate the survival signature, it is still computationally expensive.
Moreover, if network topology varies, i.e., nodes or edges are removed from the network structure, both state-enumeration and \textcolor{black}{sampling} methods need to be re-executed.
To address these challenges, two novel network reliability analysis methods are developed in the following sections. 
We first present a new \textcolor{black}{sampling} method that adopts a new scheme named K-terminal spanning tree algorithm for efficient structure function computation within MCS framework.
Then, we extend the proposed \textcolor{black}{sampling} method to a novel \textcolor{black}{active learning method}, constructing a surrogate model that can be iteratively refined to accurately capture the relationship between network behaviors and topology structures.

\section{\textcolor{black}{Proposed sampling method for network reliability analysis}} \label{section:section3} 
In this section, we initially present the K-terminal spanning tree algorithm applied to edge failure scenarios. 
\textcolor{black}{Unlike existing connectivity check algorithms to evaluate one structure function value per component state vector, the K-terminal spanning tree algorithm allows for the efficient determination of multiple structure function values by running the algorithm once.
In addition, this versatile algorithm can be utilized for both two-terminal and K-terminal connectivity checks.}
Leveraging this algorithm, we develop a new MCS method \textcolor{black}{named MC-KST} for efficient structure function computation associated with component lifetime samples and approximation of the survival signature, thereby enabling the estimation of network reliability according to Eq. (\ref{eq:NetworkReliability}).
It is worth mentioning that the component lifetime samples we adopt describe the duration until component failure, which are not corresponding to the real time $t$ used in Eq. (\ref{eq:NetworkReliability}).
Subsequently, we generalize the proposed \textcolor{black}{MC-KST} to accommodate node failure scenarios by introducing an equivalent transformation technique.

\subsection{K-terminal spanning tree algorithm} \label{section:KterminalST}
\begin{figure}[!htb]
	\centering
	\includegraphics[scale=0.83,trim=20 550 20 20,clip] {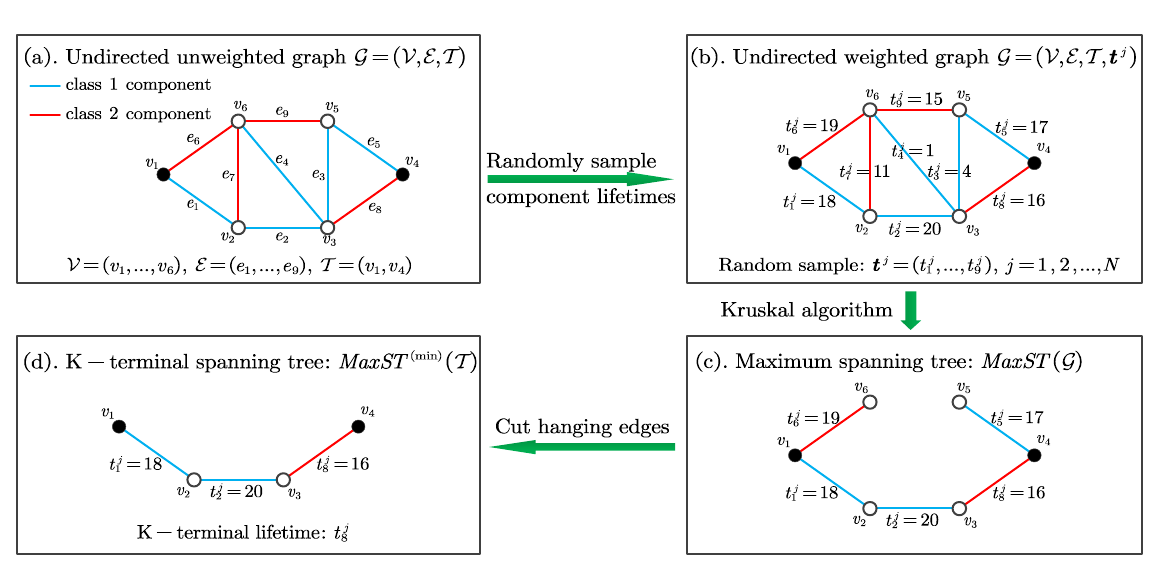}
	\caption{Basic procedure of K-terminal spanning tree algorithm}
	\label{fig:KterminalSpanningTree}
\end{figure}

Consider only edges are prone to failure while all nodes are perfectly robust. 
The basic procedure of the K-terminal spanning tree algorithm is illustrated in Figure \ref{fig:KterminalSpanningTree}, where a six-node nine-edge two-terminal network described by an undirected unweighted graph $\mathcal{G} =\left( \mathcal{V} , \mathcal{E} , \mathcal{T} \right)$ is taken as an example.
\textcolor{black}{The same procedure applies to K-terminal networks as well.}
Suppose we have a set of $N$ edge lifetime samples, denoted as $\mathbb{T} =\left( \boldsymbol{t}^1,\boldsymbol{t}^2,...,\boldsymbol{t}^{N} \right) ^\mathrm{T}$, where $\boldsymbol{t}^j=\left( t_{1}^{j}, t_{2}^{j}, ... ,t_{\mathcal{M}}^{j} \right), j=1,2,...,N$ and $t_{i}^{j}, i=1,...,\mathcal{M}$ represents the duration until the $i$-th edge fails.
The elements in the $j$-th edge lifetime sample are served as weights to generate an undirected weighted graph   
$\mathcal{G} =\left( \mathcal{V} , \mathcal{E} , \mathcal{T}, \boldsymbol{t}^j \right)$, as shown in Figure \ref{fig:KterminalSpanningTree}(b).
Then, a maximum spanning tree that connects \textcolor{black}{two terminal nodes}, denoted as $MaxST\left( \mathcal{G} \right) $ and depicted in Figure \ref{fig:KterminalSpanningTree}(c), is derived by Kruskal's algorithm \cite{kruskal1956SpanningTree}, briefly summarized in Algorithm \ref{alg:Kruskal}.
After that, the hanging edges of $MaxST\left( \mathcal{G} \right) $ are cut to obtain the K-terminal spanning tree, as Figure \ref{fig:KterminalSpanningTree}(d) presents.
This is because the hanging edges have no effect on the connection between \textcolor{black}{two terminal nodes.} 
Note that the connection is lost as soon as the edge with the shortest lifetime in the K-terminal spanning tree fails.
Hence, such shortest edge lifetime can be regarded as the most important lifetime in the K-terminal spanning tree, and is called the K-terminal lifetime (e.g., $t^j_8$ is the K-terminal lifetime in Figure \ref{fig:KterminalSpanningTree}(d)). 
\textcolor{black}{The K-terminal lifetime can be then used to determine multiple structure function values efficiently.}
Since we have $N$ edge lifetime samples, a set of $N$ K-terminal lifetimes can be obtained, which are represented as $\boldsymbol{t}_{\Bbbk}=\left( t_{\Bbbk}^{1}, t_{\Bbbk}^{2}, ... ,t_{\Bbbk}^{N} \right) ^{\mathrm{T}}$.

\begin{algorithm}[!htb]
	\caption{Maximum spanning tree by Kruskal's algorithm \cite{kruskal1956SpanningTree}}\label{alg:Kruskal}
	\begin{algorithmic}[1]
		\Require Undirected weighted graph $\mathcal{G} =\left( \mathcal{V} , \mathcal{E} , \mathcal{T}, \boldsymbol{t}^j \right), j=1,2,...,N$
		\Ensure Maximum spanning tree $MaxST \left( \mathcal{G} \right)$
		\State Sort all edges by their weights $\boldsymbol{t}^j=\left( t_{1}^{j}, t_{2}^{j}, ... ,t_{\mathcal{M}}^{j} \right)$ in decreasing order
		\State Let index $i \gets 0$ and $MaxST \left( \mathcal{G} \right) \gets \oslash$
		\While{$i < n_e$}
		\State Choose the next edge $e_{i+1}$ from the \textcolor{black}{sorted} edge set $\mathcal{E}$
		\If{adding $e_{i+1}$ to  $MaxST \left( \mathcal{G} \right)$ creates a cycle}
		\State Go to line 4
		\EndIf
		\State $MaxST\left( \mathcal{G} \right) \cup e_{i+1}$
		\State $i = i+1$
		\EndWhile
	\end{algorithmic}
\end{algorithm}

\subsection{Monte Carlo simulation method based on K-terminal spanning tree}
Leveraging the K-terminal spanning tree algorithm described above, \textcolor{black}{the MC-KST} method is established herein to evaluate the network reliability.
Initially, a set of $N_\mathrm{MCS}$ edge lifetime samples, denoted as $\mathbb{T} =\left( \boldsymbol{t}^1,\boldsymbol{t}^2,...,\boldsymbol{t}^{N_\mathrm{MCS}} \right) ^\mathrm{T}$, is generated by the MCS technique based on the CDF of each edge lifetime.
Subsequently, each edge lifetime sample $\boldsymbol{t}^j=\left( t_{1}^{j}, t_{2}^{j}, ... ,t_{\mathcal{M}}^{j} \right), j=1,2,...,{N_\mathrm{MCS}}$ is sorted in ascending order, forming an assembly denoted by $\boldsymbol{\hat{t}}=\left( \boldsymbol{\hat{t}}^1,\boldsymbol{\hat{t}}^2,...,\boldsymbol{\hat{t}}^{N_\mathrm{MCS}} \right) ^{\mathrm{T}}$.
Based on the sorted lifetime samples, a set of component state vectors is easily generated such that each $\boldsymbol{\hat{t}}^j, j=1,2,...,{N_\mathrm{MCS}}$ corresponds to $\mathcal{M}+1$ component state vectors represented by $\boldsymbol{\hat{X}}^j=\left( \boldsymbol{\hat{X}}_{1}^{j},\boldsymbol{\hat{X}}_{2}^{j},...,\boldsymbol{\hat{X}}_{\mathcal{M} +1}^{j} \right) ^{\mathrm{T}}$, where $\boldsymbol{\hat{X}}_{i}^{j}, i=1,...,\mathcal{M}+1$ is sequentially determined based on the ordering of lifetimes.
For easy understanding, consider the example depicted in Figure \ref{fig:KterminalSpanningTree}.
During the time interval $t \in \left( 0,t_4^j \right)$, all edges are operational and hence $\boldsymbol{\hat{X}}_{1}^{j}=\boldsymbol{1}$, where $\boldsymbol{1}$ denotes all-one row vector.
Similarly, during $t \in \left(t_4^j,t_5^j\right)$, only the edge $e_4$ fails, yielding $\boldsymbol{\hat{X}}_{1}^{j}=\left( 1,1,1,0,1,1,1,1,1 \right)$. This process continues until $\boldsymbol{\hat{X}}_{10}^{j}=\boldsymbol{0}$ is obtained.
The final set of component state vectors corresponding to all $N_\mathrm{MCS}$ edge lifetime samples is denoted as $S_{\boldsymbol{\hat{X}}}=\left\{ \boldsymbol{\hat{X}}^1,...,\boldsymbol{\hat{X}}^{N_\mathrm{MCS}} \right\}$, with a size of $\left( \mathcal{M} +1 \right) \times \mathcal{M} \times N_\mathrm{MCS}$.

\begin{figure}[!htb]
	\centering
	\includegraphics[scale=0.83,trim=20 485 20 20,clip] {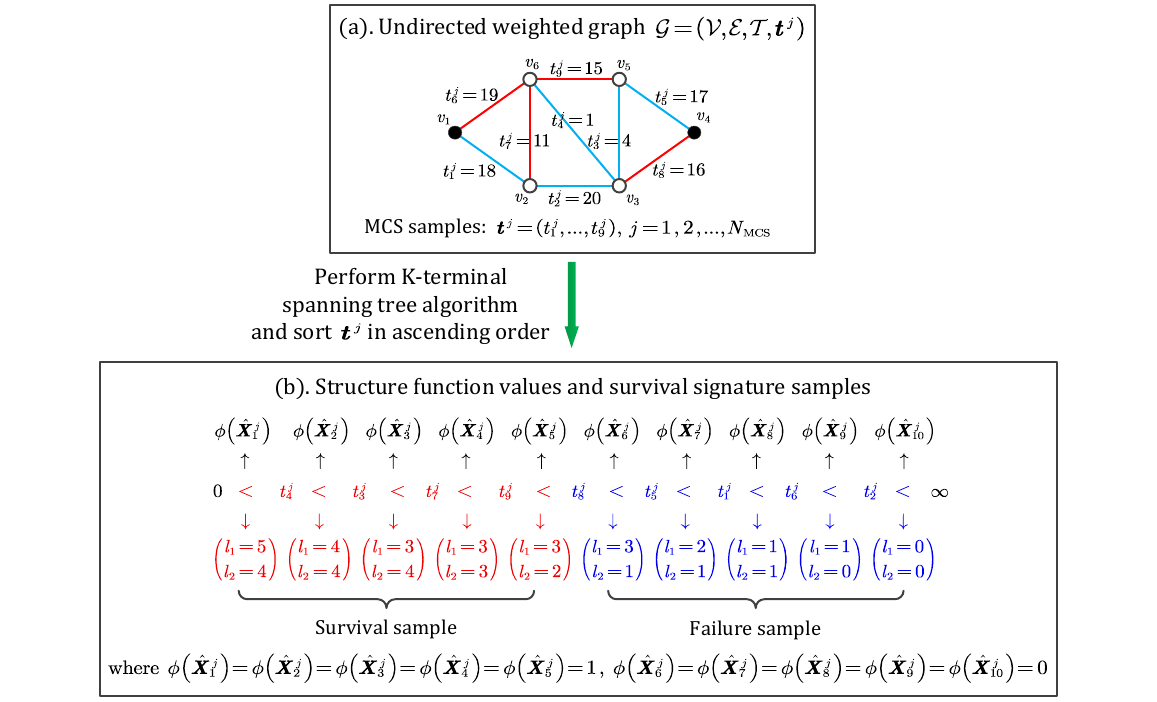}
	\caption{Description of structure function values and survival signature samples generated by the proposed MC-KST method}
	\label{fig:MCS_KterminalSpanningTree}
\end{figure}

Afterwards, the K-terminal spanning tree algorithm is performed for the lifetime sample set $\mathbb{T}$ to obtain the set of K-terminal lifetimes, i.e., $\boldsymbol{t}_{\Bbbk}=\left( t_{\Bbbk}^{1}, t_{\Bbbk}^{2}, ... ,t_{\Bbbk}^{N_{\mathrm{MCS}}} \right) ^{\mathrm{T}}$.
It is important to mention that the K-terminal lifetime determines the operating state duration of network: if the \textcolor{black}{component lifetime} is less than the K-terminal lifetime, the network functions; if the \textcolor{black}{component lifetime} exceeds the K-terminal lifetime, the network fails.
Therefore, by comparing the sorted lifetime samples with the corresponding K-terminal lifetime samples, the structure function values for $S_{\boldsymbol{\hat{X}}}$ can be determined directly without additional computational cost. 
Specifically, if the $i$-th edge lifetime in the $j$-th lifetime sample is less than the corresponding K-terminal lifetime, i.e., $t_{i}^{j}<t_{\Bbbk}^j$, the structure function value of $\boldsymbol{\hat{X}}_i ^j$ is equal to 1, i.e., $\phi \left( \boldsymbol{\hat{X}}_i ^j \right) =1$.
Otherwise, if $t_{i}^{j}\geqslant t_{\Bbbk}^j$, $\phi \left( \boldsymbol{\hat{X}}_i ^j \right) =0$.
For example, in Figure \ref{fig:MCS_KterminalSpanningTree}(b), we have $\phi \left( \boldsymbol{\hat{X}}_{1}^{j} \right) =\phi \left( \boldsymbol{\hat{X}}_{2}^{j} \right) =\phi \left( \boldsymbol{\hat{X}}_{3}^{j} \right) =\phi \left( \boldsymbol{\hat{X}}_{4}^{j} \right) =\phi \left( \boldsymbol{\hat{X}}_{5}^{j} \right) =1$ and $\phi \left( \boldsymbol{\hat{X}}_{6}^{j} \right) =\phi \left( \boldsymbol{\hat{X}}_{7}^{j} \right) =\phi \left( \boldsymbol{\hat{X}}_{8}^{j} \right) =\phi \left( \boldsymbol{\hat{X}}_{9}^{j} \right) =\phi \left( \boldsymbol{\hat{X}}_{10}^{j} \right) =0$.
Consequently, for $N_\mathrm{MCS}$ component lifetime samples, there exist $\left(\mathcal{M}+1\right) \times {N_\mathrm{MCS}}$ structure function values, whose assembly is represented by $S_{\boldsymbol{\phi}} = \left( \boldsymbol{\phi }^1,..., \boldsymbol{\phi }^{N_\mathrm{MCS}}\right)$, where $\boldsymbol{\phi }^j=\left( \phi \left( \boldsymbol{\hat{X}}_{1}^{j} \right) ,...,\phi \left( \boldsymbol{\hat{X}}_{\mathcal{M} +1}^{j} \right) \right)^\mathrm{T}, j=1,...,N_\mathrm{MCS} $.

The survival signature is then approximated based on obtained $S_{\boldsymbol{\hat{X}}}$ and $S_{\boldsymbol{\phi}}$.
As seen in Figure \ref{fig:MCS_KterminalSpanningTree}(b), for each lifetime sample, by counting the number of working components in each component class, we can identify the survival signature combinations corresponding to the component state vectors.
These survival signature combinations are classified as either survival samples or failure samples according to the values of structure function.
Specifically, if the structure function value of a component state vector is 1 (i.e., $\phi \left( \boldsymbol{\hat{X}}_{i}^{j} \right)=1, i=1,...,\mathcal{M}+1, j=1,...,{N_\mathrm{MCS}}$), it indicates that the network is still functioning, and therefore its corresponding survival signature combination obtains a survival sample; whereas if the structure function value is 0 (i.e., $\phi \left( \boldsymbol{\hat{X}}_{i}^{j} \right)=0$), the network has failed and the corresponding survival signature combination obtains a failure sample.
By performing similar classification for all component state vectors in $S_{\boldsymbol{\hat{X}}}$, a complete set of survival signature combination samples can be obtained.
Accordingly, the survival signature of each combination can be estimated by 
\begin{equation}\label{eq:Kterminal_survivalSignature}
	\Phi \left( l_1,l_2,...,l_{\mathcal{S}} \right) =\frac{N_{\mathrm{surv}}^{\left( l_1,...,l_{\mathcal{S}} \right)}}{N_{\mathrm{surv}}^{\left( l_1,...,l_{\mathcal{S}} \right)}+N_{\mathrm{fail}}^{\left( l_1,...,l_{\mathcal{S}} \right)}},
\end{equation}
where $N_{\mathrm{surv}}^{\left( l_1,...,l_{\mathcal{S}} \right)}$ and $N_{\mathrm{fail}}^{\left( l_1,...,l_{\mathcal{S}} \right)}$ denote the numbers of survival and failure samples collected based on $S_{\boldsymbol{\hat{X}}}$ and $S_{\boldsymbol{\phi}}$, respectively. 

\begin{figure}[!htb]
	\centering
	\tikzstyle{startstop} = [rectangle,rounded corners, minimum width=1.2cm, minimum height=0.8cm, text centered, draw=black, fill=red!20]
	\tikzstyle{io} = [trapezium, trapezium left angle = 70, trapezium right angle=110, minimum width=3cm, minimum height=1cm, text centered, draw=black, fill=blue!20]
	\tikzstyle{process} = [rectangle, minimum width=3cm, minimum height=0.8cm, text badly centered, draw=blue, fill=blue!15]
	\tikzstyle{decision} = [diamond,minimum width=1cm,minimum height=0.05cm,text centered,draw=green,fill=green!20,aspect=4]
	\tikzstyle{arrow} = [thick,->,>=stealth]
	\small
	
	\begin{tikzpicture}[node distance=1.4cm]
		\node (start) [startstop] {Start};
		\node (process1) [process,below of=start,align=center] {Generate edge lifetime samples $\boldsymbol{t}^j=\left( t_{1}^{j}, ... ,t_{\mathcal{M}}^{j} \right), j=1,...,{N_\mathrm{MCS}}$ by MCS};
		\node (process2) [process,below of=process1,yshift=-0.3cm,align=center] {Sort lifetimes in each $\boldsymbol{t}^j$ in ascending order. \\ Obtain component state vectors $\boldsymbol{\hat{X}}^j=\left( \boldsymbol{\hat{X}}_{1}^{j},...,\boldsymbol{\hat{X}}_{\mathcal{M} +1}^{j} \right) ^{\mathrm{T}}$ based on sorted lifetime samples $\hat{\boldsymbol{t}}^j$};
		\node (process3) [process,below of=process2,align=center,yshift=-0.6cm] {Perform K-terminal spanning tree algorithm upon each $\boldsymbol{t}^j$ \\ to attain the set of K-terminal lifetimes $\boldsymbol{t}_{\Bbbk}=\left( t_{\Bbbk}^{1}, ... ,t_{\Bbbk}^{N_{\mathrm{MCS}}} \right) ^{\mathrm{T}}$
		};
		\node (process4) [process,below of=process3,align=center,yshift=-0.3cm] {Compare each $\hat{\boldsymbol{t}}^j$ and $t_{\Bbbk}^{j}, j=1,...,N_{\mathrm{MCS}}$ to determine structure function value set $\boldsymbol{\phi }^j=\left( \phi \left( \boldsymbol{\hat{X}}_{1}^{j} \right) ,...,\phi \left( \boldsymbol{\hat{X}}_{\mathcal{M} +1}^{j} \right) \right)^\mathrm{T}$};
		\node (process5) [process,below of=process4,yshift=-0cm,align=center] {Based on $\boldsymbol{\hat{X}}^j$ and $\boldsymbol{\phi }^j$, obtain survival and failure samples of each survival signature combination};
		\node (process6) [process, below of=process5, yshift=-0.25cm,align=center] {Count the numbers of survival and failure samples for each survival signature combination, \\ and approximate survival signature by Eq. (\ref{eq:Kterminal_survivalSignature})};
		\node (process7) [process, below of=process6,yshift=-0.3cm,align=center] {Estimate network reliability by Eq. (\ref{eq:NetworkReliability})
		};
		\node (stop) [startstop, below of=process7] {Stop};
				
		\draw [arrow] (start) -- (process1);
		\draw [arrow] (process1) -- (process2);
		\draw [arrow] (process2) -- (process3);
		\draw [arrow] (process3) -- (process4);
		\draw [arrow] (process4) -- (process5);
		\draw [arrow] (process5) -- (process6);
		\draw [arrow] (process6) -- (process7);
		\draw [arrow] (process7) -- (stop);
	\end{tikzpicture}
	\caption{Flowchart of the proposed \textcolor{black}{MC-KST} method}
	\label{fig:flowchart_MCS_KterminalST}
\end{figure}
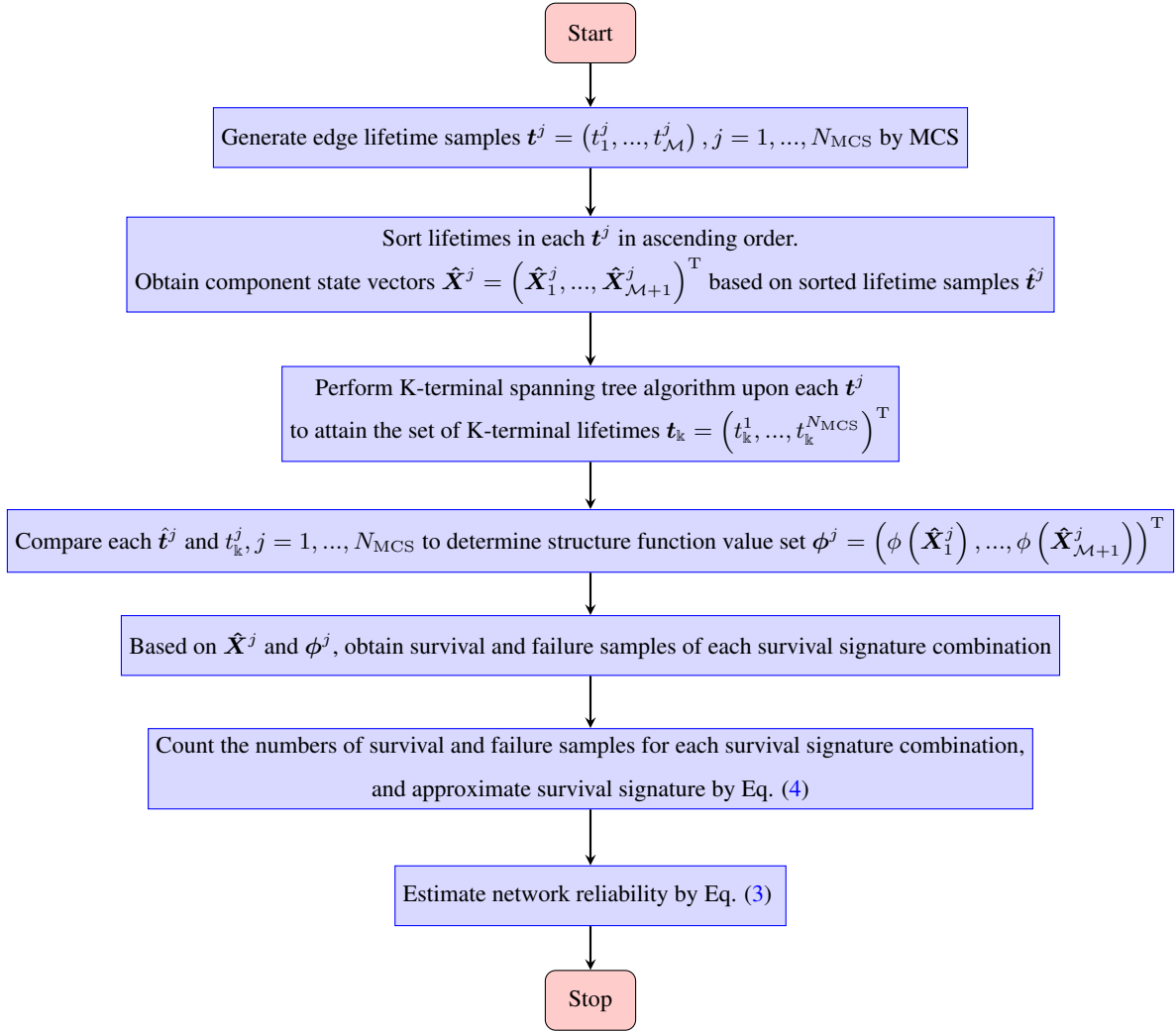

Once we obtain the survival signature, the network reliability can be estimated by Eq. (\ref{eq:NetworkReliability}).
For ease of understanding, a flowchart summarizing the procedure of the above established \textcolor{black}{MC-KST} method is shown in Figure \ref{fig:flowchart_MCS_KterminalST}.
    ~\\	

\subsection{The established \textcolor{black}{MC-KST} method for node failure problem}\label{section:edge_to_node}
Since the K-terminal spanning tree algorithm is initially established for the edge failure problem, the proposed \textcolor{black}{MC-KST} cannot be directly applied to the node failure problem.
An alternative way to address the node failure problem is to transform the problem equivalently into an edge failure problem.
It is found that for the node failure problem, if a node fails, the surrounding edges connected to the failed node also fail.
Since an edge connects two nodes, the lifetime of the edge depends on the shortest lifetime of the nodes on both sides.
Suppose we have $N$ node lifetime samples, and let the node lifetime in the $j$-th lifetime sample be $t_{v_k}^j, k=1,...,n_v, j=1,...,N$.
The equivalent transformation from node failure to edge failure can be expressed by 
\begin{equation}\label{eq: Node_to_edge}
	\tilde{t}_{e_i}^j=\min\left\{t_{v_q}^j,t_{v_p}^j  \right\},
\end{equation}
where $\tilde{t}_{e_i}^j, i=1,...,n_e$ is the edge lifetime that equivalently transformed from $t_{v_i}^j$; $v_q$ and $v_p$ denote the two nodes that connect the edge $e_i$.
Moreover, the component class of the edge is the same as the component class of the node with the shortest lifetime among the two connected nodes. 

\begin{figure}[!htb]
 	\centering
 	\includegraphics[scale=0.85,trim=20 657 30 17,clip] {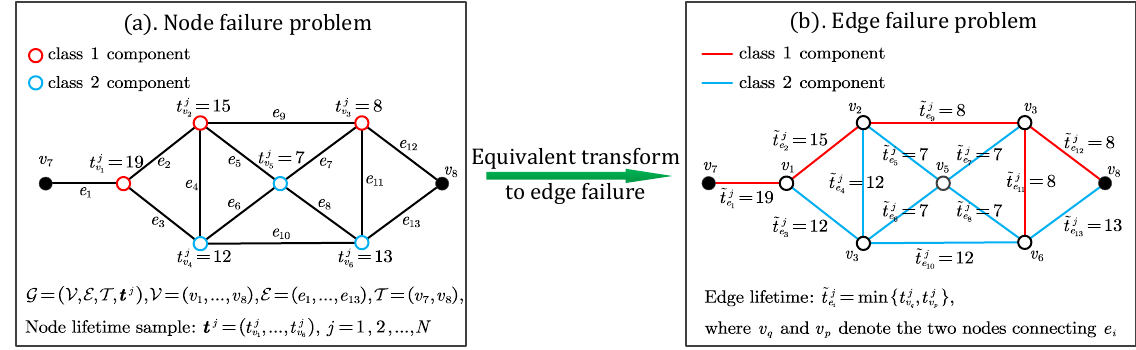}
 	\caption{Illustration of equivalent transforming node failure problem to edge failure problem}
 	\label{fig:NodeFailure_MCSKterminalSpanningTree}
 \end{figure}

To facilitate comprehension, Figure \ref{fig:NodeFailure_MCSKterminalSpanningTree} gives an example of the equivalent transformation from the node failure problem to edge failure problem. In this figure, a two-terminal network with eight nodes and thirteen edges is facing a possible node failure problem (as seen in Figure \ref{fig:NodeFailure_MCSKterminalSpanningTree}(a)).
According to Eq. (\ref{eq: Node_to_edge}), the node failure problem is equivalently transformed into an edge failure problem, where each edge has an equivalent lifetime and component class (as Figure \ref{fig:NodeFailure_MCSKterminalSpanningTree}(b) shows).
After the transformation, the \textcolor{black}{MC-KST} can be directly applied to the transformed edge failure problem for network reliability analysis. 

\section{Proposed \textcolor{black}{active learning method} for network reliability analysis}\label{section:adaptive_learning }
The proposed \textcolor{black}{MC-KST} method enables the generation of a set of component state vectors, structure function values, and survival signature combination samples through a single execution of the K-terminal spanning tree algorithm on a lifetime sample. 
 \textcolor{black}{Compared to existing MCS-based approaches, MC-KST offers notable computational efficiency enhancements by eliminating the need to compute structure function values for each component state vector separately and evaluate survival signature entries independently.} 
However, despite these improvements, ensuring accurate survival signature approximation still requires a substantial number of MCS lifetime samples. 
Moreover, because \textcolor{black}{the MC-KST} does not capture the intricate relationship between different network topology structures and network behaviors, it must be re-executed when the network topology changes. 
To address these challenges and achieve adaptability across variant networks, we develope a novel \textcolor{black}{active learning method named AL-KST, an extension of MC-KST. 
In AL-KST, we construct a surrogate model to learn network behaviors across different topology structures, enabling network response prediction without incurring additional computational costs.} 
Since computing the structure function essentially involves a classification problem because of its binary outcomes, we choose the random forest (RF) classifier as the surrogate model. 
Then, \textcolor{black}{an efficient learning function} is established to sequentially update the RF classifier, improving accuracy in structure function predictions. 
Once trained, the RF classifier enables the estimation of survival signatures for each combination based on predicted structure function values and corresponding component state vectors, \textcolor{black}{thereby} facilitating network reliability estimation. 
In the following sections, we provide detailed insights into the RF classifier and the \textcolor{black}{proposed AL-KST method}.

\subsection{Random forest classifier}
\textcolor{black}{The RF classifier is an ensemble learning method that integrates multiple base decision tree classifiers to make a final decision through a majority vote among individual trees. 
Typically, the RF adopts the classification and regression tree (CART) algorithm to train each decision tree \cite{BreiFrieStonOlsh84,breiman2001random}. 
CART begins with a root node containing all training data, and then recursively splits nodes into two, aiming to increase data homogeneity within each subset. 
This process continues until a stopping criterion is met, resulting in a leaf node that returns the majority class of training data as the final prediction \cite{wei2015comprehensive,hatwell2020chirps}. 
To reduce overfitting of individual decision trees, the RF classifier employs bootstrap sampling to create diverse subsets of the training data. 
Each subset trains a different decision tree, thereby reducing the correlation between trees and mitigating the risk of overfitting. 
The final prediction of the RF classifier aggregates the predictions of all individual trees through a majority vote, leading to a more robust and accurate prediction \cite{feng2021implementing}.} 

\begin{figure}[!htb]
	\centering
	\includegraphics[scale=0.8,trim=50 420 80 45,clip] {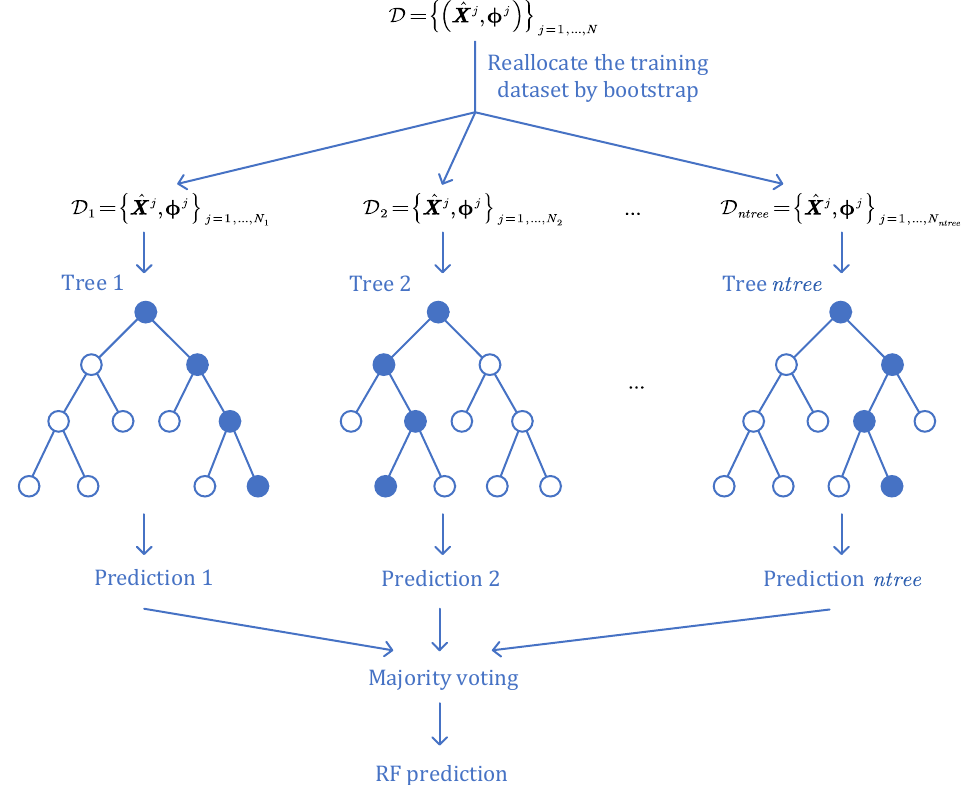}
	\caption{Illustration of random forest classifier}
	\label{fig:RandomForestClassification}
\end{figure}

\textcolor{black}{Building the RF classifier involves several steps, as illustrated in Figure \ref{fig:RandomForestClassification}. 
Let $\mathcal{D} =\left\{ \left( \boldsymbol{\hat{X}}^j,\mathbf{\phi }^j \right) \right\} _{j=1,...,N}$ denote a training dataset with $N$ samples, and $ntree$ be the number of decision trees. 
First, $ntree$ subsets of training dataset $\mathcal{D}$, denoted by $\mathcal{D} _h=\left\{ \boldsymbol{\hat{X}}^j,\mathbf{\phi }^j \right\} _{j=1,...,N_h;h=1,...,ntree}$, are produced by bootstrap sampling. 
Next, $ntree$ decision trees are trained using these subsets, following the CART algorithm. 
Finally, the predictions of all trees are combined to form the final prediction via a majority vote. 
The performance of the RF classifier can be tuned using two primary hyperparameters: $ntree$ and the number of predictors considered for each split $mtry$. 
These hyperparameters influence prediction accuracy and computational efficiency, and thus cross-validation is recommended to determine the optimal values of these hyperparameters. 
For more comprehensive details, interested readers may refer to Ref. \cite{wei2015comprehensive}.} 

\textcolor{black}{Compared to other machine-learning-based models such as support vector machines or artificial neural networks, the RF classifier offers several advantages. 
It requires less intensive hyperparameter tuning and demonstrates greater robustness to overfitting. 
Additionally, it provides measures of prediction uncertainty, which can enhance understanding of prediction confidence. 
In our context, the RF classifier enables to effectively capture the complex relationships and interactions within network topologies, making reliable predictions about network behaviors.
This characteristic makes it well-suited as a surrogate model for predicting structure function values from component state vectors.}

\subsection{\textcolor{black}{Active learning method via} random forest classifier}
Training a RF classifier with a randomly selected limited set of training samples may result in lower prediction accuracy.
Hence, we introduce \textcolor{black}{the AL-KST method} to selectively incorporate informative training samples for updating and strengthening the RF classifier during learning process.
Satisfactory prediction accuracy and robustness of the RF classifier can be achieved when a stopping criteria is met.  
The well-trained RF classifier can then be applied in subsequent network reliability analysis.

First, a pool of $N_\mathrm{MCS}$ component lifetime samples is generated by the MCS technique, denoted as $\mathbb{T} =\left( \boldsymbol{t}^1,\boldsymbol{t}^2,...,\boldsymbol{t}^{N_{\mathrm{MCS}}} \right)$ and $\boldsymbol{t}^j=\left( t_{1}^{j}, ... ,t_{\mathcal{M}}^{j} \right), j=1,...,{N_\mathrm{MCS}}$.
The initial training dataset is constructed by applying the proposed \textcolor{black}{MC-KST} method described in section \ref{section:section3} to the first $N_\mathrm{ini} = 2 \cdot \mathcal{M}$ component lifetime samples in the pool $\mathbb{T}$.
Specifically, the component lifetimes in each $\boldsymbol{t}^j, j =1,...,N_\mathrm{ini}$ are sorted in ascending order, resulting in a set of component state vectors denoted by $S_{\boldsymbol{\hat{X}},\mathrm{ini}}=\left\{ \boldsymbol{\hat{X}}^1,...,\boldsymbol{\hat{X}}^{N_{\mathrm{ini}}} \right\}$, where $\boldsymbol{\hat{X}}^j=\left( \boldsymbol{\hat{X}}_{1}^{j},...,\boldsymbol{\hat{X}}_{\mathcal{M} +1}^{j} \right) ^{\mathrm{T}}, j=1,...,N_\mathrm{ini}$.
The K-terminal spanning tree algorithm is employed upon these $N_\mathrm{ini}$ lifetime samples to generate $N_\mathrm{ini}$ K-terminal lifetimes. 
By comparing sorted lifetimes and corresponding K-terminal lifetimes, the set of structure function values for $S_{\boldsymbol{\hat{X}},\mathrm{ini}}$ is obtained, which is denoted as $S_{\boldsymbol{\phi },{\mathrm{ini}}}=\left( \boldsymbol{\phi }^1,...,\boldsymbol{\phi }^{N_\mathrm{ini}} \right)^\mathrm{T}$ with $\boldsymbol{\phi }^j=\left( \phi \left( \boldsymbol{\hat{X}}_{1}^{j} \right) ,...,\phi \left( \boldsymbol{\hat{X}}_{\mathcal{M} +1}^{j} \right) \right)^\mathrm{T}, j=1,...,N_\mathrm{ini} $.
Note that the set $S_{\boldsymbol{\hat{X}},\mathrm{ini}}$ may contain repeated component state vectors, such as $\boldsymbol{\hat{X}}_{1}^{j}=\mathbf{1}$ and $\boldsymbol{\hat{X}}_{\mathcal{M} +1}^{j}=\mathbf{0}$ for $j=1,...,N_\mathrm{ini}$.
To mitigate redundancy, the initial training dataset is then formed by combining the unique component state vectors in $S_{\boldsymbol{\hat{X}},\mathrm{ini}}$ with their corresponding $S_{\boldsymbol{\phi },{\mathrm{ini}}}$, denoted as $\mathcal{D}_\mathrm{train}=\left\{ S_{\boldsymbol{\hat{X}},\mathrm{ini},\mathrm{uniq}},S_{\boldsymbol{\phi },\mathrm{ini},\mathrm{uniq}} \right\}$.
This ensures that each unique component state vector is represented only once in the training dataset.
Meanwhile, the remaining $N_\mathrm{pred}=N_\mathrm{MCS}-N_\mathrm{ini}$ component lifetime samples in the pool $\mathbb{T}$ also undergo sorting, resulting in a set of component state vectors for prediction represented by $S_{\boldsymbol{\hat{X}},\mathrm{pred}}=\left\{ \boldsymbol{\hat{X}}^1,...,\boldsymbol{\hat{X}}^{N_{\mathrm{pred}}} \right\}$.

Based on the training dataset $\mathcal{D}_\mathrm{train}$, an RF classifier is constructed.
Using this surrogate model, the predicted structure function values of $S_{\boldsymbol{\hat{X}},\mathrm{pred}}$ are obtained, along with the prediction probability for each potential binary outcome (0 or 1).
The prediction probability is computed by the percentage of the predicted outcome among all the decision tree outcomes of the random forest.
For the $i$-th component state vector in the $j$-th lifetime sample to be predicted, i.e., $\hat{\mathbf{X}}^j_i, i=1,...,\mathcal{M}+1; j=1,...,N_\mathrm{pred}$, the prediction probabilities for the two possible binary outcomes are denoted by $\rho _{i}^{j}=P\left\{ \phi \left( \boldsymbol{\hat{X}}_{i}^{j} \right) =0 \right\}$ and $\varrho _{i}^{j}=P\left\{ \phi \left( \boldsymbol{\hat{X}}_{i}^{j} \right) =1 \right\}$, respectively. 
These probabilities indicate the uncertainty of the current RF classifier in predicting the structure function value for a new component state vector, where $\rho _{i}^{j}+\varrho _{i}^{j}=1$.
For example, if $\rho _{i}^{j}$ approaches 0 and $\varrho _{i}^{j}$ approaches 1, it signifies high confidence of the current model in predicting $\phi \left(\hat{\mathbf{X}}^j_i \right)=1$.
Conversely, if both $\rho _{i}^{j}$ and $\varrho _{i}^{j}$ hover around 0.5 with $\rho _{i}^{j}$ slightly higher, the current model is quite uncertain about predicting $\phi \left(\hat{\mathbf{X}}^j_i \right) = 0$.
The level of prediction uncertainty of model across all component state vectors in $S_{\hat{\boldsymbol{X}},\mathrm{pred}}$ can be reflected by the number of $\rho^j_i$ values falling within the range of 0.4 to 0.6, denoted by $\mathcal{N}_{\boldsymbol{\rho}}$, such that
\begin{equation}\label{eq:NumPredProbability}
	\mathcal{N}_{\boldsymbol{\rho}}=\sum_{j=1}^{N_\mathrm{pred}}{\sum_{i=1}^{\mathcal{M}+1}{I \left( \rho^j_i \in \left[ 0.4,0.6\right] \right)}},
\end{equation} 
where $I \left(\cdot\right)$ is the indicator function which returns 1 if  $\rho^j_i \in \left[ 0.4,0.6\right]$ and 0 otherwise.
The range $\left[ 0.4,0.6\right]$ is chosen because probabilities close to 0.5 reflects high uncertainty in model predictions.
Note that the higher the prediction uncertainty of RF classifier for $S_{\hat{\boldsymbol{X}},\mathrm{pred}}$, the greater instability the model predictions become, thereby limiting overall accuracy of the results.
Therefore, a stopping criterion is proposed based on the prediction uncertainty of model to determine when to stop the learning process of RF classifier.  
The stopping criterion is defined by  
\begin{equation}\label{eq:StopCriterion}
	\mathcal{N}_{\boldsymbol{\rho}}/\left({N_\mathrm{pred} \cdot \left(\mathcal{M}+1\right)}\right) \leqslant \delta, 
\end{equation}
where $\delta$ is a user-defined threshold controlling the prediction stability of the RF classifier.
It is suggested to set $\delta$ to a small value, e.g., $\delta=0.005$.

If the stopping criterion in Eq. (\ref{eq:StopCriterion}) is not satisfied, a learning function is proposed to update the RF classifier.
This learning function is developed based on minimizing the information entropy of each component lifetime sample to be predicted $\hat{\boldsymbol{t}}^j, j=1,...,N_\mathrm{pred}$, which is computed associated with the prediction probability, i.e.,
\begin{equation}
	E_{\boldsymbol{t}}^{j}=\sum_{i=1}^{\mathcal{M} +1}{\left( \rho _{i}^{j}\cdot \log \left( \rho _{i}^{j} \right) +\varrho _{i}^{j}\cdot \log \left( \varrho _{i}^{j} \right) \right)}.
\end{equation} 
This is because the information entropy measures the overall uncertainty associated with model predictions. 
Minimizing the information entropy allows the model prediction to be more stable and reliable, resulting in improving the prediction performance of the RF classifier.
In addition, considering that the set of component state vectors corresponding to a new component lifetime sample may partially overlap with the existing training dataset $\mathcal{D}_\mathrm{train}$, a weight parameter is introduced to reduce the number of duplicate vectors corresponding to potentially selected sample, i.e.,
\begin{equation}
	w^j = 1/{\mathcal{M}_\mathrm{repeat}},
\end{equation}
where ${\mathcal{M}_\mathrm{repeat}}$ denotes the number of component state vectors corresponding to $\boldsymbol{t}^{j}$ that are duplicated with $\mathcal{D}_\mathrm{train}$.
The proposed learning function aims to select samples with the highest information entropy, while also prioritizing samples with less repetition with the component state vectors in $\mathcal{D}_\mathrm{train}$.
Consequently, the learning function becomes
\begin{equation}\label{eq:learningFunction}
	\boldsymbol{t}_{\mathrm{add}}=\mathrm{arg}\max_{\boldsymbol{t}} \left\{ w^j\cdot E_{\boldsymbol{t},\mathrm{uniq}}^{j} \right\} , j=1,2,...,N_{\mathrm{pred}},
\end{equation}
where $E_{\boldsymbol{t},\mathrm{uniq}}^{j}=\sum_{i=1}^{\mathcal{M} +1-{\mathcal{M}_\mathrm{repeat}}}{\left( \rho _{i}^{j}\cdot \log \left( \rho _{i}^{j} \right) +\varrho _{i}^{j}\cdot \log \left( \varrho _{i}^{j} \right) \right)}$ is the information entropy for all unique component state vectors related to $\boldsymbol{t}^{j}$.
To further speed up the learning process and enable the use of parallel computing technique, a batch of $N_{\mathrm{add}}$ component lifetime samples are selected from the pool $\mathbb{T}$ with training samples removed according to Eq. (\ref{eq:learningFunction}). 
These $N_{\mathrm{add}}$ samples are the ones with the largest weighted information entropy values.
The set of the component state vectors related to $\boldsymbol{t}_{\mathrm{add}}$ are denoted as  $S_{\boldsymbol{\hat{X}},\mathrm{add}}=\left\{ \boldsymbol{\mathring{X}}^1,...,\boldsymbol{\mathring{X}}^{N_{\mathrm{add}}} \right\}$, where $\boldsymbol{\mathring{X}}^h=\left( \boldsymbol{\hat{X}}_{1}^{h},...,\boldsymbol{\hat{X}}_{\mathcal{M} +1}^{h} \right) ^{\mathrm{T}},h=1,...,N_{\mathrm{add}}$.
The structure function values of $S_{\boldsymbol{\hat{X}},\mathrm{add}}$ are determined according to the proposed MC-KST method, the set of which is denoted by $S_{\boldsymbol{\phi },\mathrm{add}}=\left\{ \boldsymbol{\mathring{\phi}}^1,...,\boldsymbol{\mathring{\phi}}^{N_{\mathrm{add}}} \right\}$, where $\boldsymbol{\mathring{\phi}}^h=\left( \phi \left( \boldsymbol{\hat{X}}_{1}^{h} \right) ,...,\phi \left( \boldsymbol{\hat{X}}_{\mathcal{M} +1}^{h} \right) \right) ^{\mathrm{T}},h=1,...,N_{\mathrm{add}}$.
Both $\boldsymbol{t}_{\mathrm{add}}$ and $S_{\boldsymbol{\phi },\mathrm{add}}$ are added to the training dataset $\mathcal{D}_\mathrm{train}$.
Moreover, $S_{\boldsymbol{\hat{X}},\mathrm{add}}$ are removed from the prediction set $S_{\boldsymbol{\hat{X}},\mathrm{pred}}$. 
Both of the renewed $\mathcal{D}_\mathrm{train}$ and prediction set $S_{\boldsymbol{\hat{X}},\mathrm{pred}}$ are used to update the RF classifier during learning process.

Once the stopping criterion is met, the learning process of RF classifier stops, and the final predicted structure function values associated with  $S_{\boldsymbol{\hat{X}},\mathrm{pred}}$ are accumulated to the set denoted by $S_{\boldsymbol{\tilde{\phi}},\mathrm{pred}}=\left( \boldsymbol{\tilde{\phi}}^1,...,\boldsymbol{\tilde{\phi}}^{N_{\mathrm{pred}}} \right) ^{\mathrm{T}}$, where $\boldsymbol{\tilde{\phi}}^j=\left( \tilde{\phi}\left( \boldsymbol{\hat{X}}_{1}^{j} \right) ,...,\tilde{\phi}\left( \boldsymbol{\hat{X}}_{\mathcal{M} +1}^{j} \right) \right) ^{\mathrm{T}},j=1,...,N_{\mathrm{pred}}$.
The final training dataset $\mathcal{D}_\mathrm{train}$ and the prediction dataset $\mathcal{D}_\mathrm{pred} = \left\{ S_{\boldsymbol{\hat{X}},\mathrm{pred}}, S_{\boldsymbol{\tilde{\phi}},\mathrm{pred}} \right\}$ are employed to obtain the survival and failure samples of each survival signature combination.
By counting the numbers of these survival and failure samples, the survival signature of each combination is then approximated by Eq. (\ref{eq:Kterminal_survivalSignature}).
Subsequently, the network reliability is evaluated by substituting the approximated survival signature into Eq. (\ref{eq:NetworkReliability}).

\subsection{Step-by-step procedure}

\begin{figure}[!htb]
    \centering
	\tikzstyle{startstop} = [rectangle,rounded corners, minimum width=1.2cm, minimum height=0.8cm, text centered, draw=black, fill=red!20]
	\tikzstyle{io} = [trapezium, trapezium left angle = 70, trapezium right angle=110, minimum width=3cm, minimum height=1cm, text centered, draw=black, fill=blue!20]
	\tikzstyle{process} = [rectangle, minimum width=3cm, minimum height=0.8cm, text badly centered, draw=blue, fill=blue!15]
	\tikzstyle{decision} = [diamond,minimum width=1cm,minimum height=0.05cm,text centered,draw=green,fill=green!20,aspect=4]
	\tikzstyle{arrow} = [thick,->,>=stealth]
	\small
	
	\begin{tikzpicture}[node distance=1.4cm]
	\node (start) [startstop] {Start};
	\node (process1) [process,below of=start,align=center] {Set ${N} _{\mathrm{MCS}}$, $ntree$, $mtry$ and $\delta$, and let $N=N_\mathrm{ini}$};
	\node (process2) [process,below of=process1,yshift=0cm,align=center] {Generate component lifetime sample pool $\mathbb{T}$ by the MCS technique};
	\node (process3) [process,below of=process2,align=center,yshift=-0.4cm] 
         {Obtain initial training dataset $\mathcal{D} _{\mathrm{train}}=\left\{ S_{\boldsymbol{\hat{X}},\mathrm{ini},\mathrm{uniq}},S_{\boldsymbol{\phi },{\mathrm{ini},\mathrm{uniq}}} \right\}$ based on $\mathbb{T}$ and \\ the proposed \textcolor{black}{MC-KST} method, and yield prediction set $S_{\boldsymbol{\hat{X}},\mathrm{pred}}$ from $\mathbb{T}$};
	\node (process4) [process,below of=process3,align=center,yshift=-0.7cm] {Train an RF classifier with $\mathcal{D}_\mathrm{train}$ \\ and make prediction upon $S_{\boldsymbol{\hat{X}},\mathrm{pred}}$ by trained model};
	\node (decision1) [decision,below of=process4,yshift=-0.5cm] {Criterion satisfied?};
	\node (process5) [process,below of=decision1,yshift=-0.6cm,align=center] {Learn $\boldsymbol{t}_\mathrm{add}$ by learning function in Eq. (\ref{eq:learningFunction}) \\ and obtain $S_{\boldsymbol{\hat{X}},\mathrm{add}}$ associated with $\boldsymbol{t}_\mathrm{add}$};
	\node (process6) [process, below of=process5, yshift= -0.6 cm,align=center] {Compute $S_{\boldsymbol{\phi},\mathrm{add}}$ related to $S_{\boldsymbol{\hat{X}},\mathrm{add}}$ using the proposed  \textcolor{black}{MC-KST}, \\  add $S_{\boldsymbol{\hat{X}},\mathrm{add}}$ and $S_{\boldsymbol{\phi},\mathrm{add}}$ into $\mathcal{D}_\mathrm{train}$ and remove $S_{\boldsymbol{\hat{X}},\mathrm{add}}$ from $S_{\boldsymbol{\hat{X}},\mathrm{pred}}$};
    \node (process7) [process,below of=process6,yshift=-0.4cm,align=center] {$N=N+N_\mathrm{add}$};
    \node[shape=coordinate](y1) [left of = process7, xshift= -4 cm] {Y1};
	\node (process8) [process,right of=decision1,xshift=6.8cm,align=center] {Output $\mathcal{D}_\mathrm{train}$ and $\mathcal{D}_\mathrm{pred}= \left\{ S_{\boldsymbol{\hat{X}},\mathrm{pred}}, S_{\boldsymbol{\tilde{\phi}},\mathrm{pred}} \right\}$, \\and estimate the survival signature by Eq. (\ref{eq:Kterminal_survivalSignature})};
	\node (process9) [process,below of=process8,yshift=-0.65cm,align=center] {Calculate network reliability by substituting \\ the estimated survival signature into Eq. (\ref{eq:NetworkReliability})};
	\node (stop) [startstop, below of=process9,yshift=-0.6cm,align=center] {Stop};

	\draw [arrow] (start) -- (process1);
	\draw [arrow] (process1) -- (process2);
	\draw [arrow] (process2) -- (process3);
	\draw [arrow] (process3) -- (process4);
	\draw [arrow] (process4) -- (decision1);
	\draw [arrow] (decision1) -- node[anchor=east] {No} (process5);
        \draw [arrow] (process5) -- (process6);
        \draw [arrow] (process6) -- (process7);
    \draw  (process7) --  (y1);
    \draw [arrow] (y1) |- (process4);
	\draw [arrow] (decision1) -- node[above] {Yes} (process8);
	\draw [arrow] (process8) -- (process9);
	\draw [arrow] (process9) -- (stop);
	\end{tikzpicture}
	\caption{Flowchart of the proposed \textcolor{black}{AL-KST method}}
	\label{fig:flowchart2}
\end{figure}
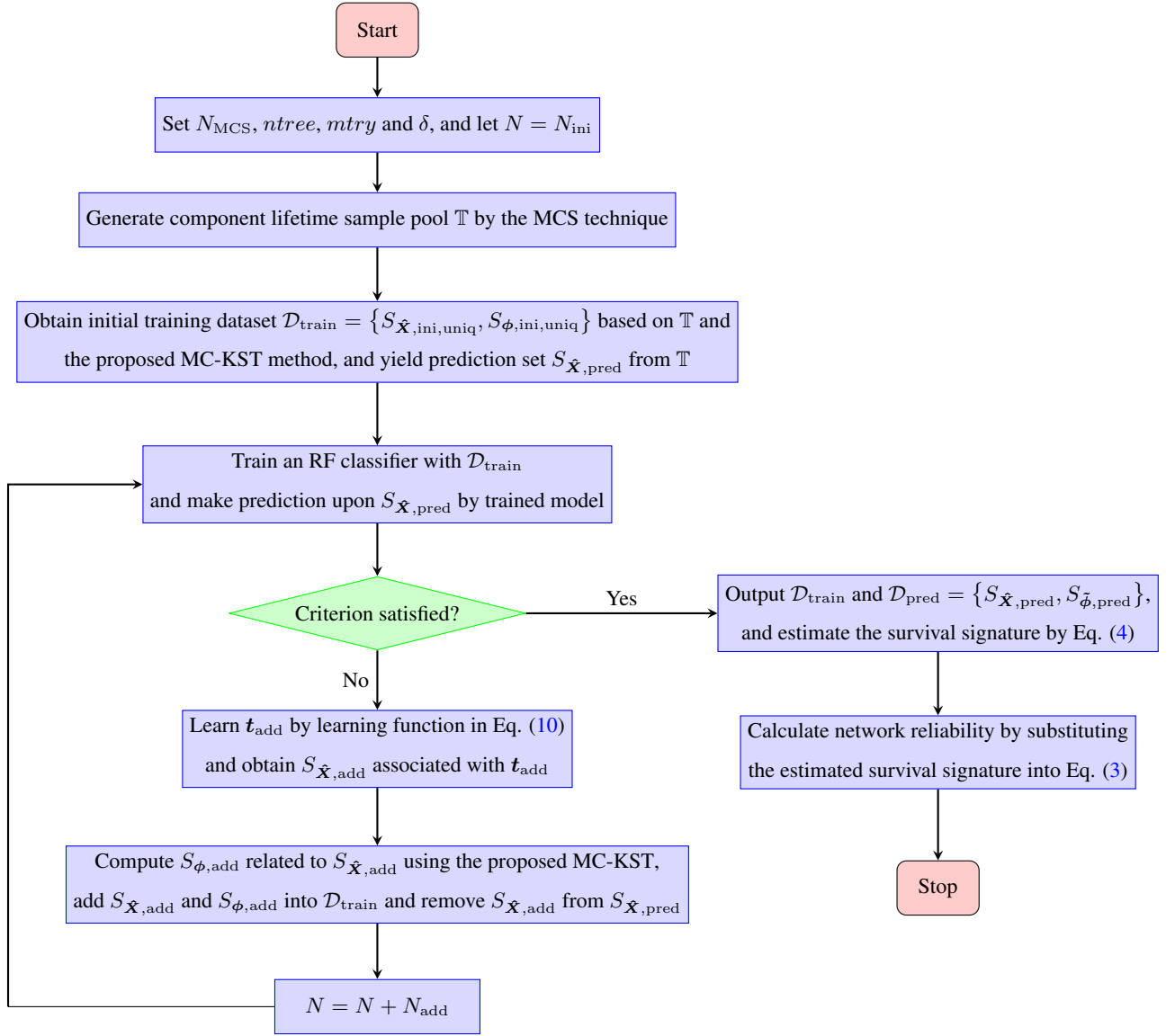

A flowchart of the \textcolor{black}{proposed AL-KST method} is shown in Fig. \ref{fig:flowchart2}.
In addition, a brief procedure of \textcolor{black}{AL-KST} is summarized as follows:
    ~\\
    
	\textbf{Step 1}: Initialization. 
	Set the size of component lifetime sample pool $N _{\mathrm{MCS}}$, the number of samples forming the initial training dataset $N_\mathrm{ini}$, the number of samples added at each iteration $N_\mathrm{add}$, the number of trees $ntree$ and the number of predictor variables for each decision split $mtry$ in RF classifier, and stopping tolerance $\delta$. 
	Create the initial training dataset $\mathcal{D} _{\mathrm{train}}$ by three steps.
	First, employ the MCS technique to generate a sample pool containing $N_\mathrm{MCS}$ component lifetime samples $\mathbb{T}=\left(\boldsymbol{t}^1,...,\boldsymbol{t}^{N_\mathrm{MCS}}\right)$.
	Then, apply the proposed \textcolor{black}{MC-KST} method to the first $N _{\mathrm{ini}}$ lifetime samples to obtain the initial set of component state vectors $S_{\boldsymbol{\hat{X}},\mathrm{ini}}=\left\{ \boldsymbol{\hat{X}}^1,...,\boldsymbol{\hat{X}}^{N_{\mathrm{ini}}} \right\}$ and the set of structure function values $S_{\boldsymbol{\phi },\mathrm{ini}}=\left( \boldsymbol{\phi }^1,...,\boldsymbol{\phi }^{N_{\mathrm{ini}}} \right) ^{\mathrm{T}}$.
	Subsequently, combine the unique vectors in $S_{\boldsymbol{\hat{X}},\mathrm{ini}}$ with their associated values in  $S_{\boldsymbol{\phi },\mathrm{ini}}$ to form the initial training dataset $\mathcal{D} _{\mathrm{train}}=\left\{ S_{\boldsymbol{\hat{X}},\mathrm{ini},\mathrm{uniq}},S_{\boldsymbol{\phi },{\mathrm{ini},\mathrm{uniq}}} \right\} $. 
	Denote the number of component lifetime samples used to build $\mathcal{D} _{\mathrm{train}}$ as $N$, where $N=N_\mathrm{ini}$ at present.
	The rest $N_\mathrm{pred}=N_\mathrm{MCS}-N_\mathrm{ini}$ sorted component lifetime samples are utilized to yield the prediction set $S_{\boldsymbol{\hat{X}},\mathrm{pred}}=\left\{ \boldsymbol{\hat{X}}^1,...,\boldsymbol{\hat{X}}^{N_{\mathrm{pred}}} \right\}$.

	\textbf{Step 2}: Construct the RF classifier.
	By employing the \textit{TreeBagger} function in the Matlab "Statistics and Machine Learning Toolbox" and hyperparameters $ntree$ and $mtry$, the RF classifier between different component state vectors and structure function values is trained based on current $\mathcal{D} _{\mathrm{train}}$.
	Then, make a prediction upon $S_{\boldsymbol{\hat{X}},\mathrm{pred}}$ using the trained RF classifier, resulting in predicted structure function values $\tilde{\phi}\left( \boldsymbol{\hat{X}}_{i}^{j} \right)$ and prediction probabilities $\left\{ \rho_i^j, \varrho_i^j \right\}$ for each $\hat{\mathbf{X}}^j_i, i=1,...,\mathcal{M}+1; j=1,...,N_\mathrm{pred}$. 

	\textbf{Step 3}: Check the stopping criterion.
	To ensure the robustness of model predictions, a delayed judgment is recommended, wherein the stopping criterion in Eq. (\ref{eq:StopCriterion}) is consecutively checked by two times.
	If the delayed judgment is met, end the learning process and go to step 5;
	otherwise, go to step 4.
	
	\textbf{Step 4}: Enrichment of the training dataset.
	A batch of $N_{\mathrm{add}}$ component lifetime samples $\boldsymbol{t}_\mathrm{add}$ are selected from $\mathbb{T}$ by maximizing the weighted information entropy in Eq. (\ref{eq:learningFunction}).
	The set of component state vectors related to $\boldsymbol{t}_\mathrm{add}$ is denoted as $S_{\boldsymbol{\hat{X}},\mathrm{add}}$.
	The structure function values of $S_{\boldsymbol{\hat{X}},\mathrm{add}}$ are determined according to the proposed MC-KST method.
	Then, $S_{\boldsymbol{\hat{X}},\mathrm{add}}$ and corresponding structure function values $S_{\boldsymbol{\phi},\mathrm{add}}$ are added to $\mathcal{D}_\mathrm{train}$, and $S_{\boldsymbol{\hat{X}},\mathrm{add}}$ are removed from $S_{\boldsymbol{\hat{X}},\mathrm{pred}}$.
	Afterwards, set $N=N+N_\mathrm{add}$ and go to Step 2 to update the RF classifier.
	
	\textbf{Step 5}: Compute the network reliability.
	This step involves first outputting the final $\mathcal{D}_\mathrm{train}$ and prediction dataset  $\mathcal{D}_\mathrm{pred} = \left\{ S_{\boldsymbol{\hat{X}},\mathrm{pred}}, S_{\boldsymbol{\tilde{\phi}},\mathrm{pred}} \right\}$, where $S_{\boldsymbol{\tilde{\phi}},\mathrm{pred}}=\left( \boldsymbol{\tilde{\phi}}^1,...,\boldsymbol{\tilde{\phi}}^{N_{\mathrm{pred}}} \right) ^{\mathrm{T}}$ and $\boldsymbol{\tilde{\phi}}^j=\left( \tilde{\phi}\left( \boldsymbol{\hat{X}}_{1}^{j} \right) ,...,\tilde{\phi}\left( \boldsymbol{\hat{X}}_{\mathcal{M} +1}^{j} \right) \right) ^{\mathrm{T}},j=1,...,N_{\mathrm{pred}}$. 
	Afterwards, based on $\mathcal{D}_\mathrm{train}$ and $\mathcal{D}_\mathrm{pred}$, the survival and failure samples for each survival signature combination are attained.
	By counting the number of survival and failure samples for each combination, the survival signature is then computed according to Eq. (\ref{eq:Kterminal_survivalSignature}).
	Finally, the network reliability is calculated by Eq. (\ref{eq:NetworkReliability}).
    ~\\	
	
\section{Test examples}\label{section:section5}
 \textcolor{black}{In this section, we explore six synthetic network cases and two real-world network applications to validate the efficacy of two proposed methods for two-terminal network reliability estimation.}
The initial training dataset $\mathcal{D}\mathrm{train}$ begins with $N\mathrm{ini}=2 \cdot n_v$ samples. 
Each iteration during learning process adds $N_{\mathrm{add}}=2(n_v-2) + 4n_e$ samples. 
Hyperparameters of the RF classifier, determined through cross-validation, are set with $ntree=100$ and $mtry=\mathcal{M}$. 
The learning process is halted when the tolerance threshold reaches $\delta=0.005$.
Analytical reliability results are obtained via an enumeration method that computes the survival signature by enumerating all $2^\mathcal{M}$ possible component state vectors and computing corresponding structure function values using Dijkstra algorithm \cite{dijkstra2022note}.
To illustrate the accuracy of the two proposed methods, the relative errors between analytical and predicted reliability results are given, which are computed by $RE\left( t \right) ={{\left| R_{\mathrm{true}}\left( t \right) -R_{\mathrm{pred}}\left( t \right) \right|}/{R_{\mathrm{true}}\left( t \right)}}$, where $R_{\mathrm{true}}\left( t \right)$ and $R_{\mathrm{pred}}\left( t \right)$ represent analytical and predicted results, respectively.
\textcolor{black}{Additionally, we present predicted results from an RF classifier trained directly using randomly extracted samples from the sample pool, without employing active learning. 
This demonstration aims to underscore the effectiveness of the proposed AL-KST. 
We refer to the RF classifier without the learning method as RF-KST in the following.
The number of samples used in RF-KST matches the total number of samples employed in AL-KST.}
All of these methods are implemented using MATLAB R2022b on the same computer equipped with an Intel Core i7-11800H processor operating at 2.30 GHz and 32GB of RAM, \textcolor{black}{utilizing 8 processors}.

\subsection{A node failure network with a variant}
\begin{figure}[!htb]
	\centering
	\subfigure[\textcolor{black}{Original network}]{
		\begin{minipage}{8.0cm}
			\centering
			\includegraphics[scale=0.9, trim=30 420 300 260,clip]{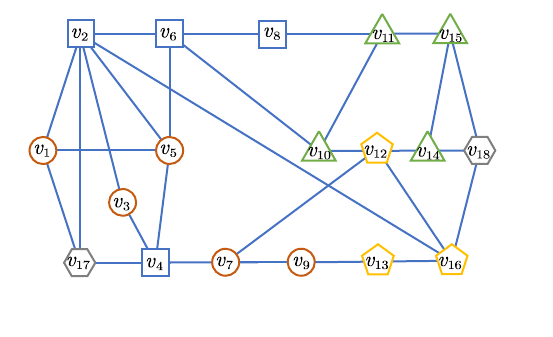}
		\end{minipage}
		\label{fig:originalNetwork}
	}%
	\subfigure[Variant network]{
		\begin{minipage}{8.0cm}
			\centering
			\includegraphics[scale=0.9,trim=300 420 25 260,clip]{example1_node_failure_network.pdf}
		\end{minipage}
		\label{fig:variantNetwork}
	}%
	\caption{Illustration of the original node failure network and its variant for example 1}
	\label{fig:example1_network}
\end{figure}

A node failure network consisting of 18 nodes and 29 edges is first investigated, which is modified from Ref. \cite{ramirez2009deterministic}. 
The original network and its variant are visually presented in Figure \ref{fig:example1_network}. 
Within the original network configuration, nodes $v_{17}$ and $v_{18}$ are designated as the source and target nodes, respectively. 
The remaining 16 nodes are categorized into four distinct classes: nodes $\left\{ v_2,v_4,v_6,v_8 \right\}$ belong to class 1, nodes $\left\{ v_1,v_5,v_7,v_9 \right\}$ belong to class 2, nodes $\left\{ v_{10},v_{11},v_{14} \right\}$ belong to class 3, and nodes $\left\{ v_{12},v_{13},v_{16} \right\}$ belong to class 4.
In Figure \ref{fig:example1_network}, the differentiation of components belonging to class 1 to 4 is visually characterized by rectangular, circle, triangular, and hexagon markers, respectively.
The lifetime distributions of components are as follows:
Components in class 1 follow the exponential distribution with an inverse scale parameter $\lambda=0.8$, class 2 components follow the weibull distribution with scale parameter $\alpha_{\mathrm{wbl}}=1.7$ and shape parameter $\beta_{\mathrm{wbl}}=3.6$, class 3 components follow the lognormal distribution with location parameter $\mu=1.5$ and scale parameter $\sigma=2.6$, and class 4 components follow the gamma distribution with scale parameter $\alpha_{\mathrm{gam}}=3.1$ and shape parameter $\beta_{\mathrm{gam}}=1.5$.
The variant network, depicted in Figure \ref{fig:variantNetwork}, is formed by removing nodes $v_{3}$ and $v_{15}$ from the original network.

\begin{figure}[!htb]
	\centering
	\subfigure[Network reliability results]{
		\begin{minipage}{8.0cm}
			\centering
			\includegraphics[scale=0.6, trim=15 0 0 0,clip]{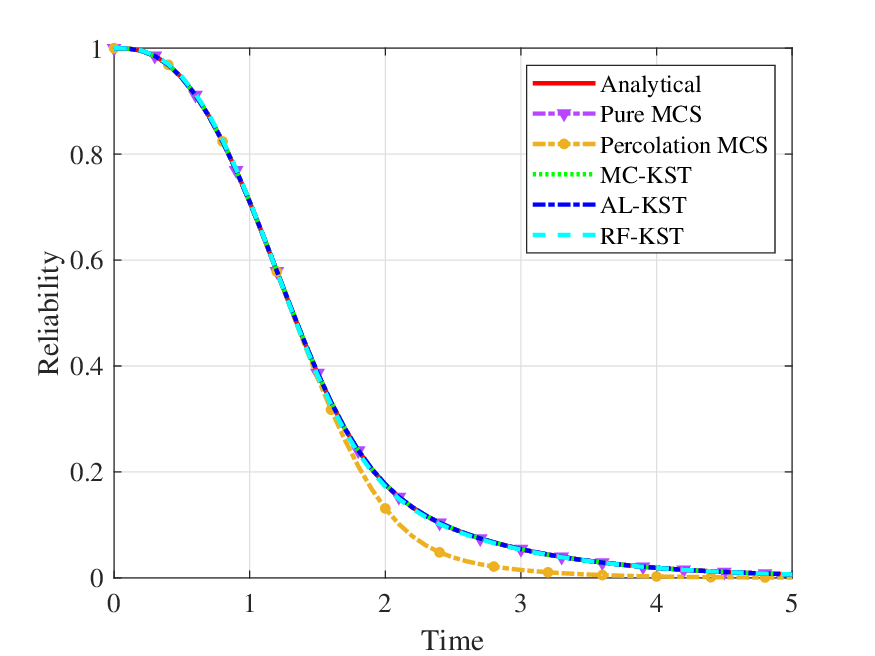}
		\end{minipage}
		\label{fig:ex1_reliability}
	}%
	\subfigure[Relative errors of network reliability]{
		\begin{minipage}{8.0cm}
			\centering
			\includegraphics[scale=0.6, trim=15 0 0 0,clip]{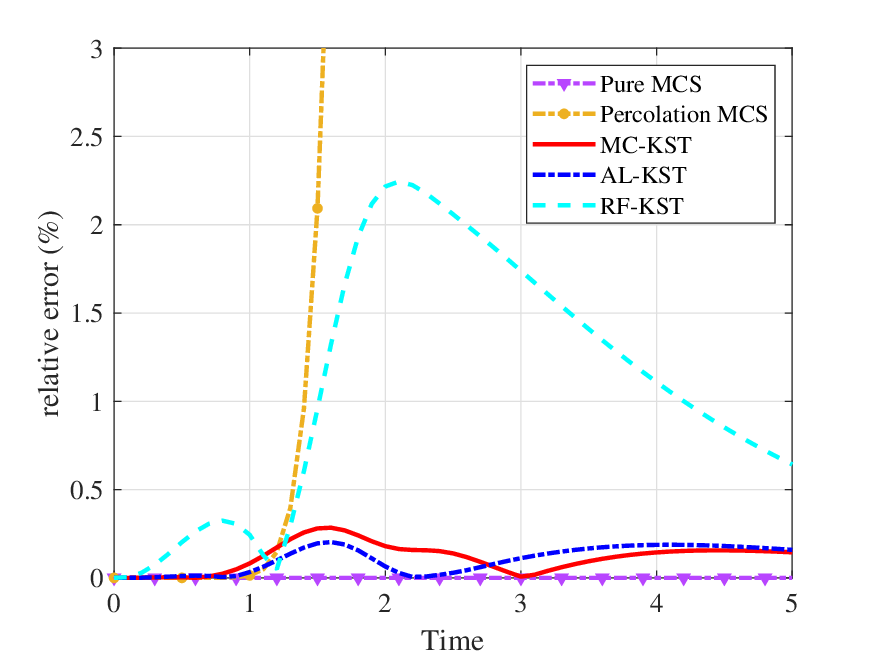}
		\end{minipage}
		\label{fig:ex1_RE_reliability}
	}%
	\caption{\textcolor{black}{Network reliability results and related relative errors by different methods for original network in example 1}}
	\label{fig:example1_result}
\end{figure}

To implement the proposed two methods for network reliability analysis in the original network shown in Figure \ref{fig:originalNetwork}, a component lifetime sample pool \textcolor{black}{of size $N_\mathrm{MCS}=10,000$} is first generated using the MCS technique according to the lifetime distribution of each component.
Note that this example is a node failure problem, and therefore requires the equivalent transformation from node lifetime samples to edge lifetime samples according to Eq. (\ref{eq: Node_to_edge}).
Denote the transformed sample pool by \textcolor{black}{$\mathbb{T} =\left( \boldsymbol{t}^1,...,\boldsymbol{t}^{N_\mathrm{MCS}} \right) ^\mathrm{T}$}.
The proposed \textcolor{black}{MC-KST} evaluates the network reliability by executing the K-terminal spanning tree algorithm \textcolor{black}{10,000} times over the entire $\mathbb{T}$, as detailed in section \ref{section:section3}.
On the other hand, the proposed \textcolor{black}{AL-KST} extracts the first  $N_\mathrm{ini}=36$ component lifetime samples from $\mathbb{T}$ to form an initial training dataset $\mathcal{D}_\mathrm{train}$, which is employed to train the RF classifier.
Each learning iteration selects $N_\mathrm{add}=152$ lifetime samples with the largest weighted entropy values from $\mathbb{T}$ to enrich $\mathcal{D}_\mathrm{train}$, refining the RF classifier accordingly.
The learning process ends after three iterations, where a total of $N_\mathrm{train}=492$ times of the K-terminal spanning tree algorithm are performed.
The network reliability results and associated relative errors by both proposed methods are illustrated in Figure \ref{fig:example1_result}, compared with the analytical result from the enumeration method.
\textcolor{black}{Besides, the computational costs and maximum relative errors of results for both proposed methods are recorded in Table \ref{tab:example1_oriNetwork_MaxRelativeError_CPUtime}.
It can be observed that the estimated results obtained by the proposed two methods closely align with the analytical result, showing very small relative errors with maximum 0.2840\% for MC-KST and 0.2027\% for AL-KST.
These negligible discrepancies underscore the accuracy of the proposed two methods in estimating network reliability for the original network shown in Figure \ref{fig:originalNetwork}.}

\begin{table}[htb!]
	\caption{\textcolor{black}{Computational costs and maximum relative errors by different methods for original network in example 1}}
	\setlength\tabcolsep{3pt}
	\label{tab:example1_oriNetwork_MaxRelativeError_CPUtime}
	\centering 	
	\begin{threeparttable}
		\begin{tabular}{llllllll}
			\toprule 
			Method & CPU time & $N$ & $RE_{\max}$ & Method & CPU time & $N$ & $RE_{\max}$ \\
			\midrule       
			Analytical     & 4.10 s & 65536   & - & MC-KST    & 1.88 s & 10000   & 0.2840\%\\
			Pure MCS     & 4.10 s & 65536   & 0\% &AL-KST    & 8.28 s & 492   & 0.2027\%\\
			Percolation MCS     & 3.75 s & 50643   & 92.6664\% & RF-KST    & 2.39 s & 492   & 2.2446\%  \\
			\bottomrule       
		\end{tabular}
		\begin{tablenotes}
			\item \textcolor{black}{Note: $RE_{\max}$ and $N$ denote the maximum relative error of reliability and the total number of samples for each result.}
		\end{tablenotes}
	\end{threeparttable}
\end{table}

\textcolor{black}{In addition, to verify the effectiveness of proposed MC-KST and AL-KST methods, we conduct a comparative analysis of network reliability against two state-of-the-art MCS-based techniques: the pure MCS from Ref. \cite{behrensdorf2018efficient} and percolation MCS from Ref. \cite{behrensdorf2021numerically}, as well as the offline surrogate-based method RF-KST.
The pure MCS method approximates each combination of survival signature by utilizing a maximum of 10,000 samples per combination. 
On the other hand, the percolation MCS method leverages percolation theory to identify and filter trivial combinations of survival signature, and then employs MCS approximation to estimate the remaining combinations with a similar limit of 10,000 MCS samples for each combination.
The RF-KST is comprised by an RF classifier trained directly with $N_\mathrm{train}=492$ samples randomly extracted from $\mathbb{T}$.
We present the network reliability results and relative errors of these comparative analyses in Figure \ref{fig:example1_result}, while summarizing computational costs, including CPU time, and maximum relative errors in Table \ref{tab:example1_oriNetwork_MaxRelativeError_CPUtime}. 
Among the methods, pure MCS offers the highest accuracy but requires more CPU time than MC-KST. 
MC-KST, on the other hand, provides a good balance by achieving lower CPU time (1.88 seconds) while maintaining acceptable accuracy (maximum relative error of 0.2840\%).
The percolation MCS method struggles to achieve accurate reliability results due to excessive filtering by the percolation theory, resulting in a maximum relative error of 92.6664\%.
The RF-KST requires less CPU time (2.39 seconds), but it exhibits higher relative errors (2.2446\%) compared to both proposed methods.
The higher CPU time for AL-KST (8.10 seconds) is attributed to the iterative update process of the random forest (RF) classifier, but this additional computation improves the accuracy of the reliability estimates, making AL-KST a robust option despite the increased CPU time.
}

\begin{figure}[!htb]
	\centering
	\subfigure[Network reliability results]{
		\begin{minipage}{8.0cm}
			\centering
			\includegraphics[scale=0.6, trim=15 0 15 0,clip]{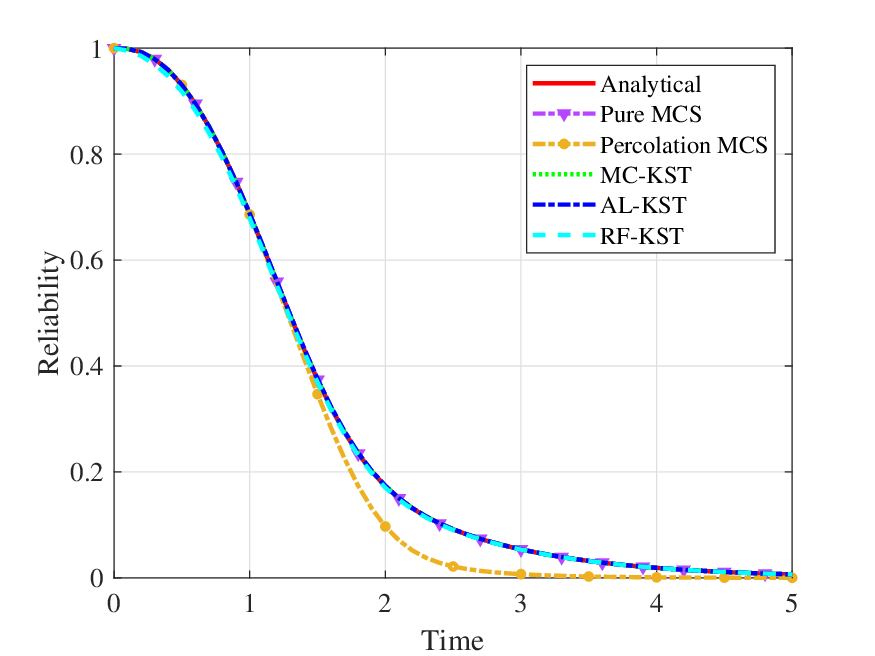}
		\end{minipage}
		\label{fig:ex1variant_reliability}
	}%
	\subfigure[Relative errors of network reliability]{
		\begin{minipage}{8.0cm}
			\centering
			\includegraphics[scale=0.6, trim=10 0 15 0,clip]{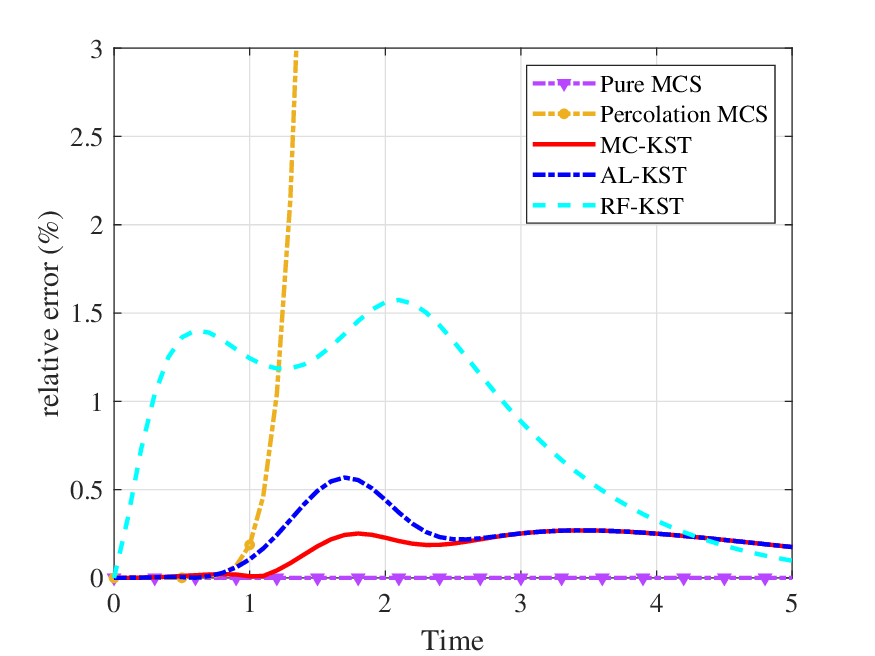}
		\end{minipage}
		\label{fig:ex1variant_RE_reliability}
	}%
	\caption{Network reliability results and related relative errors by different methods for variant network in example 1}
	\label{fig:example1_variant_result}
\end{figure}

For the variant network depicted in Figure \ref{fig:variantNetwork}, a new sample pool comprising \textcolor{black}{10000} new component lifetime samples is generated via the MCS technique to constitute the set of component state vectors pertinent to the variant network.
\textcolor{black}{To assess the network reliability for this variant network, the proposed AL-KST method can directly use the pre-trained RF classifier from the original network to predict the new set of component state vectors and estimate network reliability. 
In contrast, the proposed MC-KST must be executed again on the entire new sample pool. 
Similarly, the analytical method, pure MCS and percolation MCS need to be re-executed for the variant network. 
This demonstrates the adaptability advantage of AL-KST to different network topologies.
Figure \ref{fig:example1_variant_result} illustrates the network reliability results and corresponding relative errors predicted by both proposed methods, along with the results from pure MCS, percolation MCS, the RF-KST constructed for the original network, and the analytical result.
Evidently, both proposed methods still yield reliability results closely aligned with the analytical result and exhibit very small relative errors. 
While pure MCS provides accurate reliability results, percolation MCS fails to do so.
The reliability obtained by the RF-KST slightly deviates from the analytical result, showing a larger relative error compared to the two proposed methods. 
Table \ref{tab:example1_variantNet_MaxRelativeError_CPUtime} lists the computational costs and maximum relative errors for all these methods on the variant network. 
This table shows that AL-KST significantly outperforms other methods in terms of CPU time (0.35 seconds) while maintaining acceptable accuracy (maximum relative error of 0.5683\%).
This efficiency is due to the ability of the RF classifier, once trained for the original network, can be directly used for the variant network.
In addition, although RF-KST is also quick because of using the RF classifier, it manifests a larger maximum relative error (1.5738\%) compared with AL-KST.
The MC-KST, pure MCS, and percolation MCS methods, along with the analytical result, require longer CPU times than AL-KST and RF-KST, due to the re-execution for the variant network.
However, MC-KST remains faster than pure MCS and percolation MCS, providing accurate reliability estimates for the variant network.
These observations demonstrate that the proposed MC-KST and AL-KST methods effectively balance accuracy and computational efficiency, with AL-KST offering a more efficient alternative for variant networks due to its adaptability.}

\begin{table}[htb!]
	\caption{\textcolor{black}{Computational costs and maximum relative errors by different methods for variant network in example 1}}
	\setlength\tabcolsep{3pt}
	\label{tab:example1_variantNet_MaxRelativeError_CPUtime}
	\centering 	
		\begin{tabular}{llllllll}
			\toprule 
			Method & CPU time & $N$ & $RE_{\max}$ & Method & CPU time & $N$ & $RE_{\max}$ \\
			\midrule       
			Analytical     & 2.62 s & 16384   & - & MC-KST    & 1.62 s & 10000   & 0.2688\%\\
			Pure MCS     & 2.19 s & 16384   & 0\% &AL-KST    & 0.35 s & -   & 0.5683\%\\
			Percolation MCS     & 2.08 s & 9908   & 97.9497\% & RF-KST    & 0.33 s & 492   & 1.5738\%  \\
			\bottomrule       
		\end{tabular}
\end{table}

\subsection{An edge failure network with a variant}
\begin{figure}[!htb]
	\centering
	\subfigure[Original network]{
		\begin{minipage}{8.0cm}
			\centering
			\includegraphics[scale=1, trim=30 640 400 50,clip]{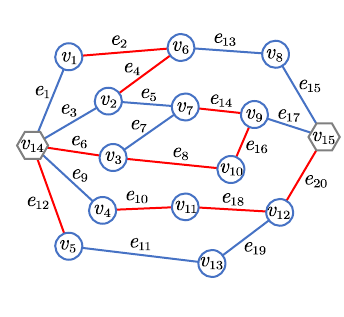}
		\end{minipage}
		\label{fig:ex2originalNetwork}
	}%
	\subfigure[Variant network]{
		\begin{minipage}{8.0cm}
			\centering
			\includegraphics[scale=1,trim=400 640 25 50,clip]{example2_edge_failure_network.pdf}
		\end{minipage}
		\label{fig:ex2variantNetwork}
	}%
	\caption{Illustration of the original edge failure network and its variant for example 2}
	\label{fig:example2_network}
\end{figure}

Example 2 explores an edge failure network with 15 nodes and 20 edges.
Figure \ref{fig:example2_network} shows the original network and its variant.
In the original network, the nodes $v_{14}$ and $v_{15}$ are designated as source and target nodes, respectively.
The 20 edges are divided into two distinct classes such that 10 first odd-numbered edges are designated as class 1 and the remaining even-numbered edges belong to class 2. 
Figure \ref{fig:ex2originalNetwork} depicts the original network, where the red line represents class 1 components governed by an exponential distribution with an inverse scale parameter $\lambda_1=0.8$, while the blue line denotes class 2 components characterized by an exponential distribution parameterized by $\lambda_2=1.5$.
The variant network, displayed in \ref{fig:ex2variantNetwork} is derived by removing four edges $\left\{e_1,e_8,e_{13},e_{16}\right\}$ from the original network.

\begin{figure}[!htb]
	\centering
	\subfigure[Network reliability results]{
		\begin{minipage}{8.0cm}
			\centering
			\includegraphics[scale=0.6, trim=15 0 0 0,clip]{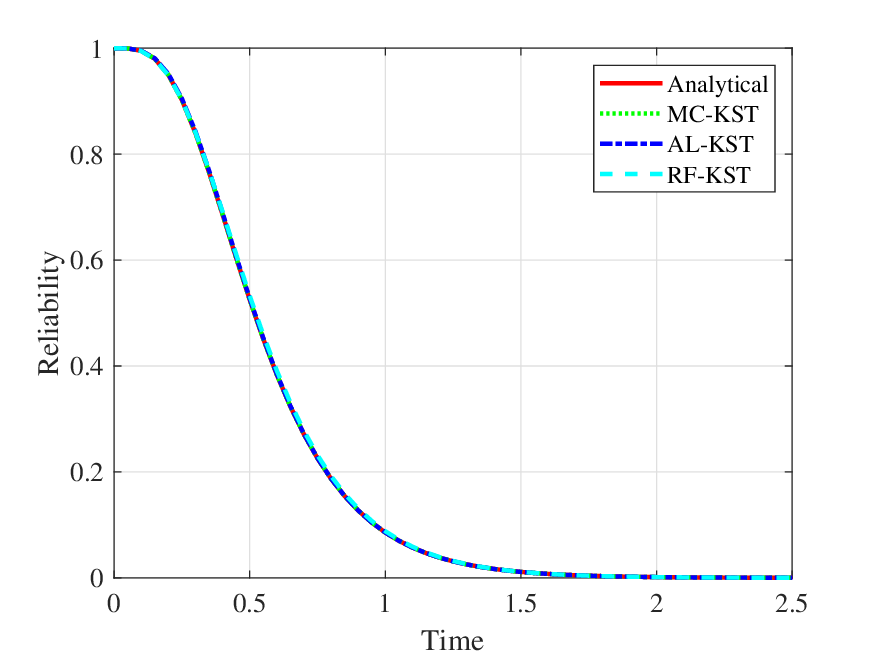}
		\end{minipage}
		\label{fig:ex2_reliability}
	}%
	\subfigure[Relative errors of network reliability]{
		\begin{minipage}{8.0cm}
			\centering
			\includegraphics[scale=0.6, trim=15 0 0 0,clip]{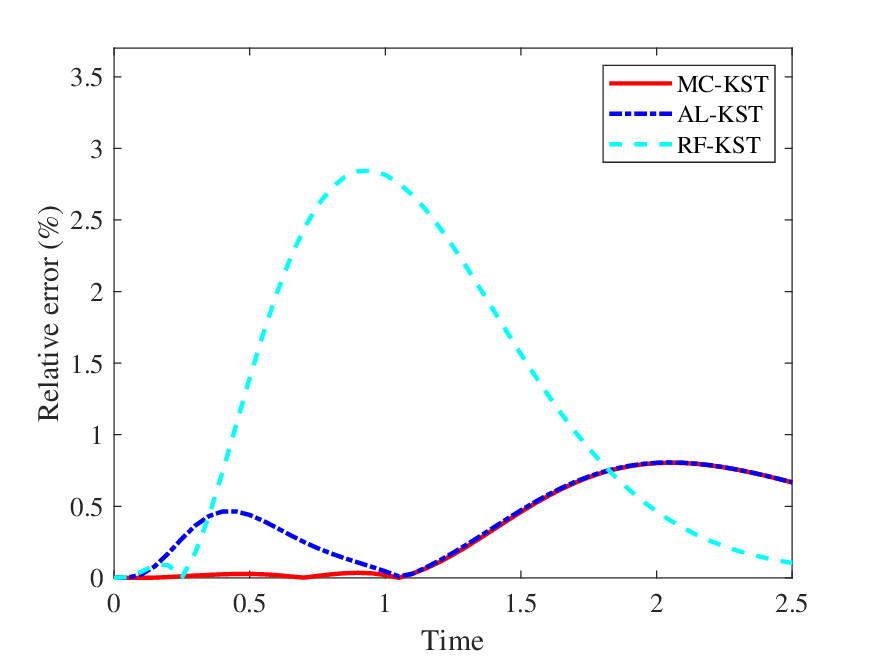}
		\end{minipage}
		\label{fig:ex2_RE_reliability}
	}%
	\caption{Network reliability results and related relative errors by different methods for original network in example 2}
	\label{fig:example2_result}
\end{figure}

In this example, the proposed \textcolor{black}{MC-KST} method adopts 50000 component lifetime samples to evaluate the network reliability concerning edge failure.
The proposed \textcolor{black}{AL-KST} begin with using $N_\mathrm{ini}=40$ component lifetime samples to train the RF classifier, and then refine the model by sequentially adding $N_\mathrm{add}=110$ component lifetime samples in each iteration.
The refinement process concludes after 9 iterations, incorporating a total of $N_\mathrm{train}=1030$ component lifetime samples and 1030 times of performing the K-terminal spanning tree algorithm.
The network reliability predictions and associated relative errors by both proposed methods are illustrated in Figure \ref{fig:example2_result}.
Additionally, the results predicted by \textcolor{black}{RF-KST with its RF classifier constructed by a random subset of 1030 samples from the sample pool}, are also given in Figure \ref{fig:example2_result}.
The results demonstrate that both proposed methods can accurately predict the reliability for the edge failure network in this example, with a maximum relative error of 0.8063\%. 
However, \textcolor{black}{AL-KST} exhibits higher computational efficiency compared to \textcolor{black}{MC-KST}. 
Moreover, \textcolor{black}{RF-KST} shows larger relative error in reliability compared to the analytical result.

\begin{figure}[!htb]
	\centering
	\subfigure[Network reliability results]{
		\begin{minipage}{8.0cm}
			\centering
			\includegraphics[scale=0.6, trim=15 0 15 0,clip]{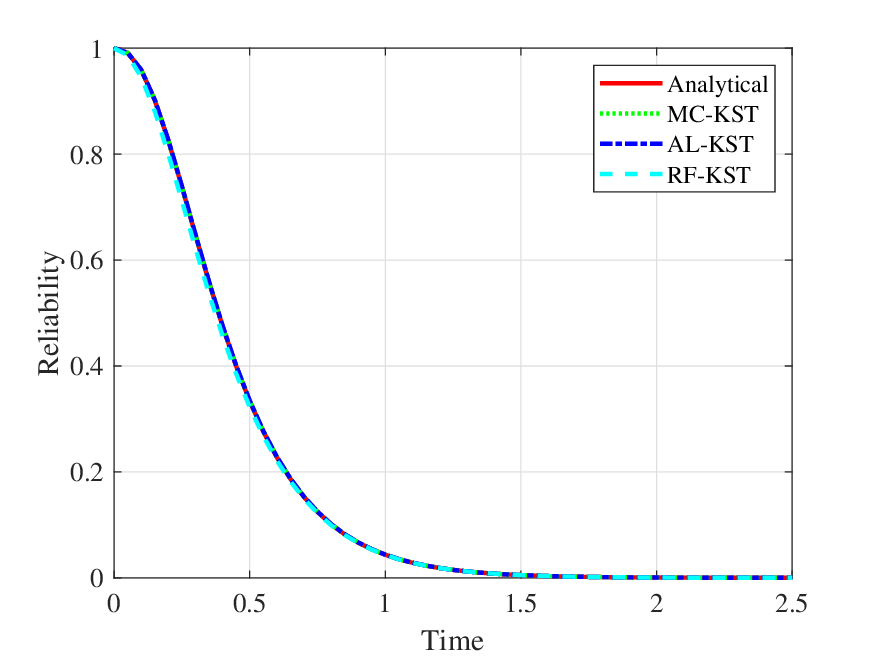}
		\end{minipage}
		\label{fig:ex2variant_reliability}
	}%
	\subfigure[Relative errors of network reliability]{
		\begin{minipage}{8.0cm}
			\centering
			\includegraphics[scale=0.6, trim=10 0 15 0,clip]{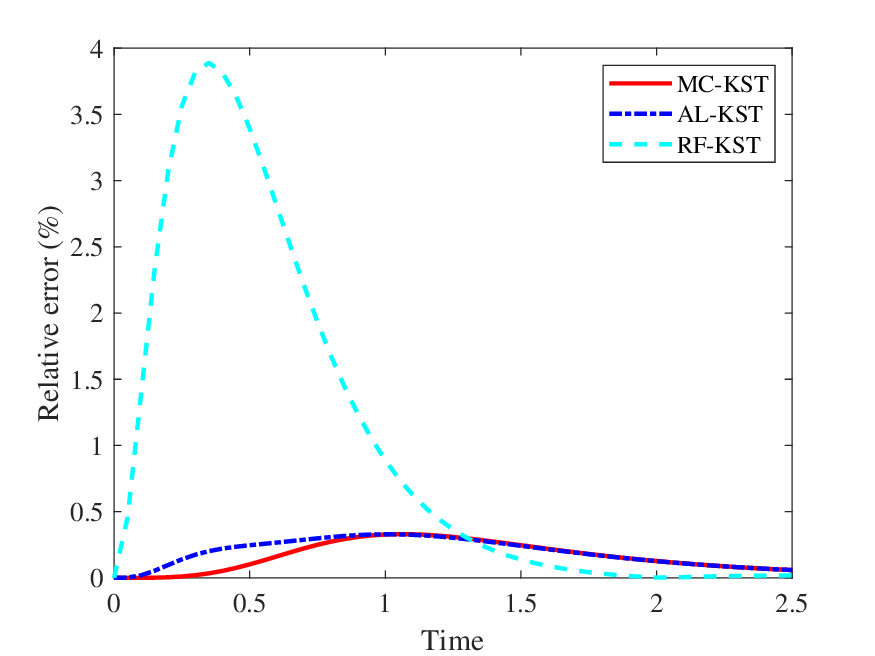}
		\end{minipage}
		\label{fig:ex2variant_RE_reliability}
	}%
	\caption{Network reliability results and related relative errors by different methods for variant network in example 2}
	\label{fig:example2_variant_result}
\end{figure}

For the variant network displayed in Figure \ref{fig:ex2variantNetwork}, the proposed \textcolor{black}{MC-KST} method is re-executed using 50000 new component lifetime samples associated with the variant network.
Conversely, the proposed \textcolor{black}{AL-KST} straightforwardly employs the RF classifier trained for the original network to predict the reliability of the variant network.
The reliability predictions and relative errors obtained from both proposed methods are shown in Figure \ref{fig:example2_variant_result}, together with the results by\textcolor{black}{RF-KST using the RF classifier} trained for the original network.
Similarly, the reliability prediction results by proposed two methods match the analyzed result quite well, with both relative errors less than 0.5\%.
Nonetheless, the prediction result by \textcolor{black}{RF-KST} deviates slightly from the analytical result with a relative error greater than 3.5\%.

\subsection{Four different networks considering node or edge failure}
\begin{figure}[!htb]
	\centering
 	\subfigure[Network 1]{
		\begin{minipage}{8.0cm}
			\centering
			\includegraphics[scale=0.9, trim=25 560 300 55,clip]{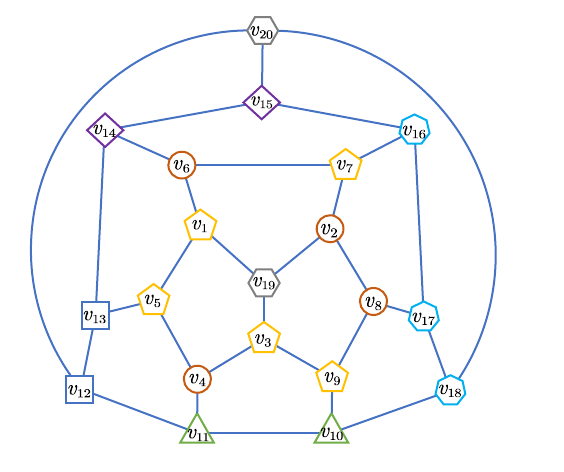}
		\end{minipage}
		\label{fig:ex3_Network1}
	}%
	\subfigure[Network 2]{
		\begin{minipage}{8.0cm}
			\centering
			\includegraphics[scale=0.9,trim=300 560 15 55,clip]{example3_four_test_network.pdf}
		\end{minipage}
		\label{fig:ex3_Network2}
	}%
 
	\subfigure[Network 3]{
		\begin{minipage}{8.0cm}
			\centering
			\includegraphics[scale=0.9,trim=25 260 300 350,clip]{example3_four_test_network.pdf}
		\end{minipage}
		\label{fig:ex3_Network3}
	}%
	\subfigure[Network 4]{
	\begin{minipage}{8.0cm}
		\centering
		\includegraphics[scale=0.9,trim=300 260 35 350,clip]{example3_four_test_network.pdf}
	\end{minipage}
	\label{fig:ex3_Network4}
}%
	\caption{Illustration of two node failure networks (network 1 and 2) and two edge failure networks (network 3 and 4)}
	\label{fig:ex3_four_network}
\end{figure}

Four different networks, as displayed in Figure \ref{fig:ex3_four_network}, are modified from Ref. \cite{ramirez2009deterministic} and Ref. \cite{reed2019efficient} to further testify the efficacy of the proposed two methods for network reliability analysis.
In all networks, nodes denoted by hexagonal markers are the source and target nodes.
The first two networks, labeled 'Network 1' and 'Network 2', concern only node failure, where components denoted by circle markers constitute class 1, components represented by pentagonal markers belong to class 2, components denoted by triangular markers are assigned to class 3, components represented by rectangular markers belong to class 4, components denoted by rhombic markers comprise class 5, and components represented by octagonal markers are assigned to class 6.
The remaining two networks, labeled 'Network 3' and 'Network 4', are two edge failure networks, where the yellow lines denote the class 1 components, the green lines signify the class 2 components, and the blue lines represent the class 3 components.
Further details regarding the components for each network can be found in Table \ref{tab:description_four_network}.

\begin{table}[htb!]
	\caption{Description of components for four different networks}
	\setlength\tabcolsep{3pt}
	\label{tab:description_four_network}
	\centering 	
	\begin{threeparttable}
	\begin{tabular}{llll}
		\toprule 
		Network name  & Type of component failure & Component class  & Statistical information   \\
		\midrule       
		\multirow{6}{*}{Network 1}     & \multirow{6}{*}{Node failure}   & 1    &Exponential(0.8) \\
		& & 2    &Exponential(1.5) \\
		& & 3    &Weibull(1.5, 3.2) \\
		& & 4    &Weibull(1.8, 3.5) \\
		& & 5    &Lognormal(1.5, 2.5) \\
		& & 6    &Gamma(3.0, 1.2) \\
		\multirow{2}{*}{Network 2}     & \multirow{2}{*}{Node failure}   & 1      & Exponential(1.2)  \\
		& & 2      & Exponential(0.7)  \\
		\multirow{3}{*}{Network 3}   & \multirow{3}{*}{Edge failure}    & 1      & Exponential(1.2)  \\
		& & 2      & Weibull(1.5, 3.6)  \\
		& & 3      & Lognormal(1.5, 2.6)  \\
		\multirow{3}{*}{Network 4}   & \multirow{3}{*}{Edge failure}   	& 1      & Exponential(1.2)  \\
		& & 2      & Weibull(1.5, 3.6)  \\
		& & 3      & Gamma(3.0, 1.8)  \\
		\bottomrule       
	\end{tabular}
	\begin{tablenotes}
		\item Note: The distribution parameters in the last column include the inverse scale parameters for exponential distribution, the scale and shape parameters for weibull distribution and gamma distribution, and the location and scale parameters for lognormal distribution. 
	\end{tablenotes}
	\end{threeparttable}
\end{table}

\begin{figure}[!htb]
	\centering
	\subfigure[Network reliability results]{
		\begin{minipage}{8.0cm}
			\centering
			\includegraphics[scale=0.6, trim=15 0 0 0,clip]{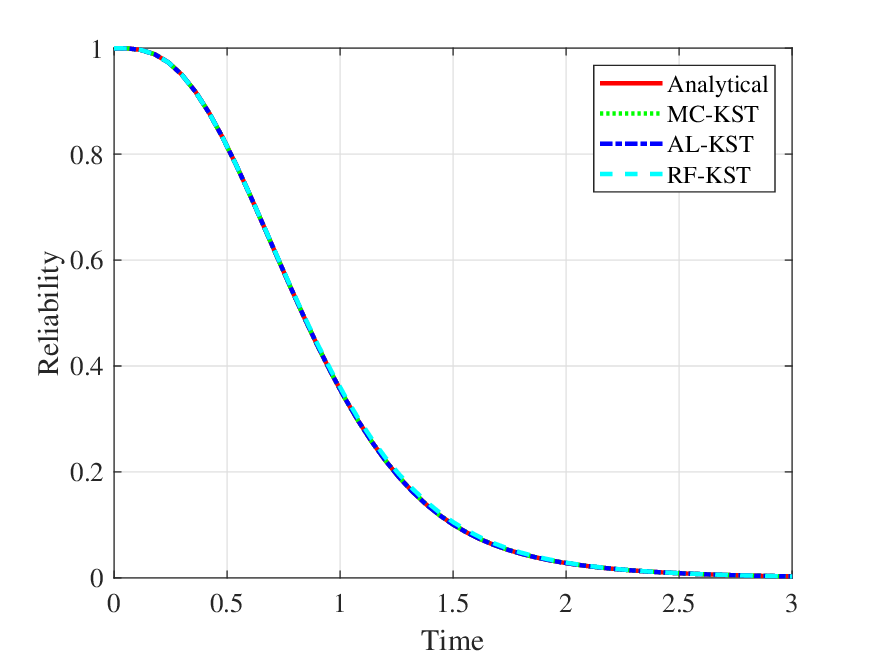}
		\end{minipage}
		\label{fig:ex3_network1_reliability}
	}%
	\subfigure[Relative errors of network reliability]{
		\begin{minipage}{8.0cm}
			\centering
			\includegraphics[scale=0.6, trim=15 0 0 0,clip]{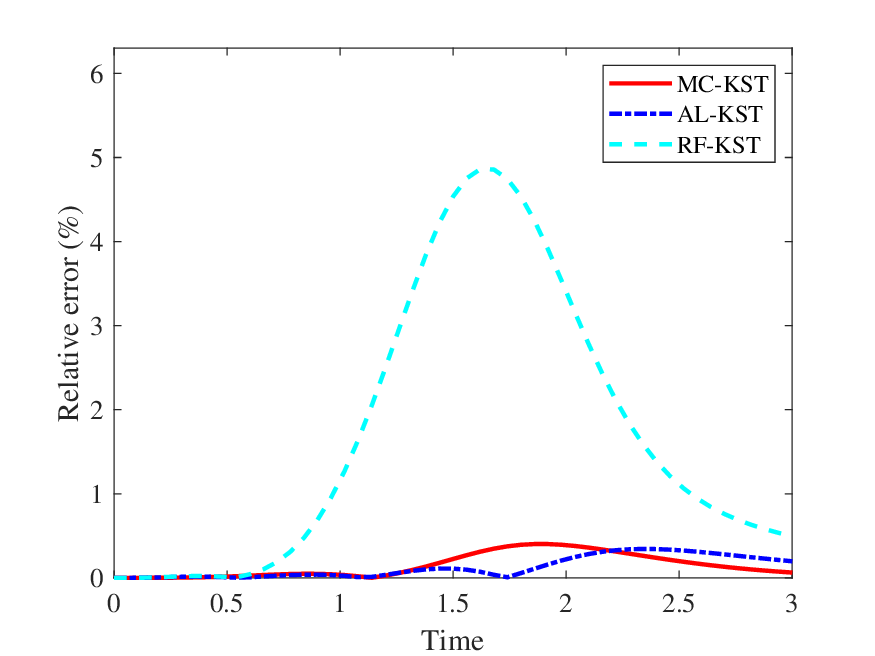}
		\end{minipage}
		\label{fig:ex3_network1_RE_reliability}
	}%
	\caption{Network reliability results and related relative errors by different methods for node failure network 1}
	\label{fig:example3_network1_result}
\end{figure}

\begin{figure}[!htb]
\centering
\subfigure[Network reliability results]{
	\begin{minipage}{8.0cm}
		\centering
		\includegraphics[scale=0.6, trim=15 0 0 0,clip]{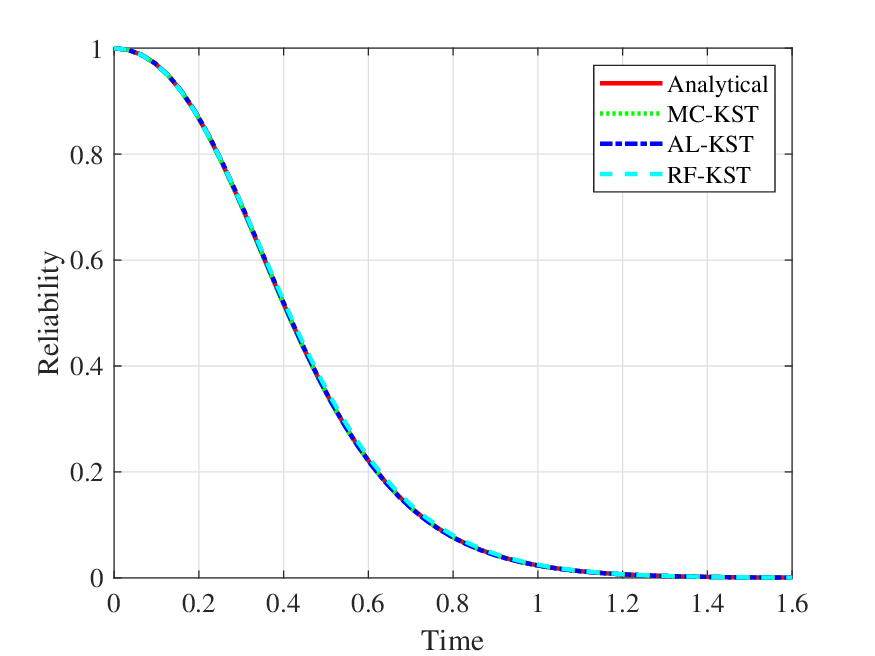}
	\end{minipage}
	\label{fig:ex3_network2_reliability}
}%
\subfigure[Relative errors of network reliability]{
	\begin{minipage}{8.0cm}
		\centering
		\includegraphics[scale=0.6, trim=15 0 0 0,clip]{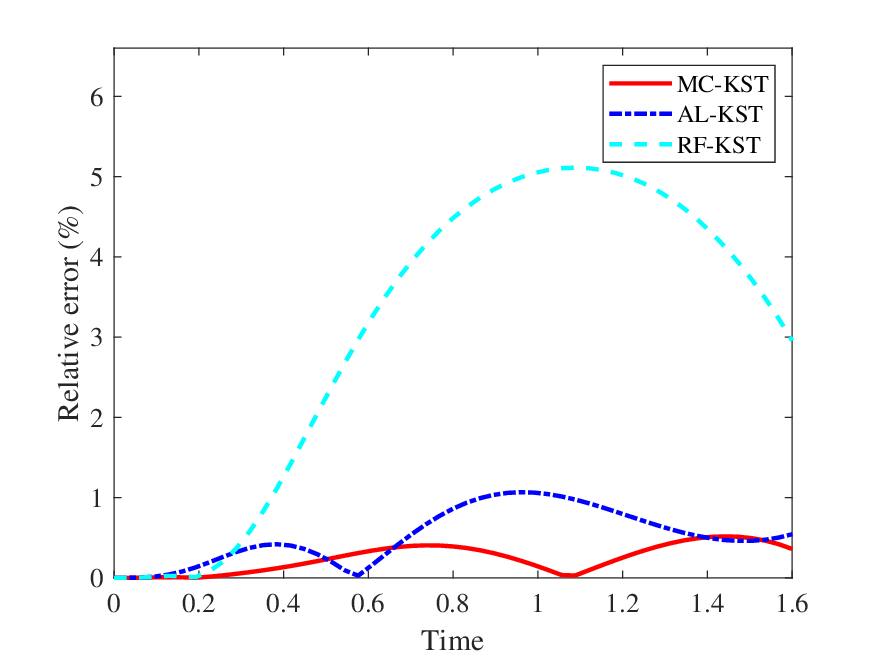}
	\end{minipage}
	\label{fig:ex3_network2_RE_reliability}
}%
\caption{Network reliability results and related relative errors by different methods for node failure network 2}
\label{fig:example3_network2_result}
\end{figure}

\begin{figure}[!htb]
\centering
\subfigure[Network reliability results]{
	\begin{minipage}{8.0cm}
		\centering
		\includegraphics[scale=0.6, trim=15 0 0 0,clip]{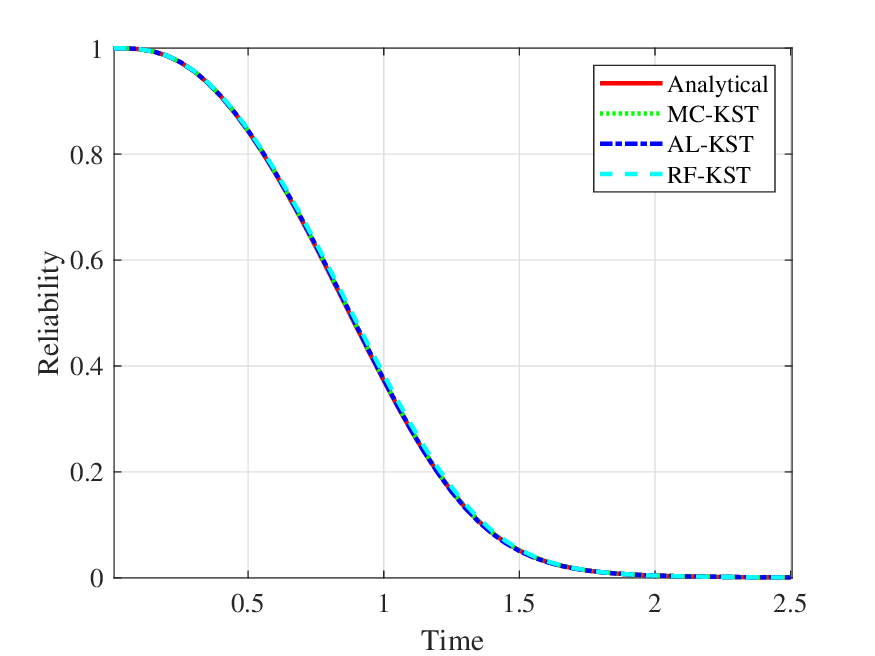}
	\end{minipage}
	\label{fig:ex3_network3_reliability}
}%
\subfigure[Relative errors of network reliability]{
	\begin{minipage}{8.0cm}
		\centering
		\includegraphics[scale=0.6, trim=15 0 0 0,clip]{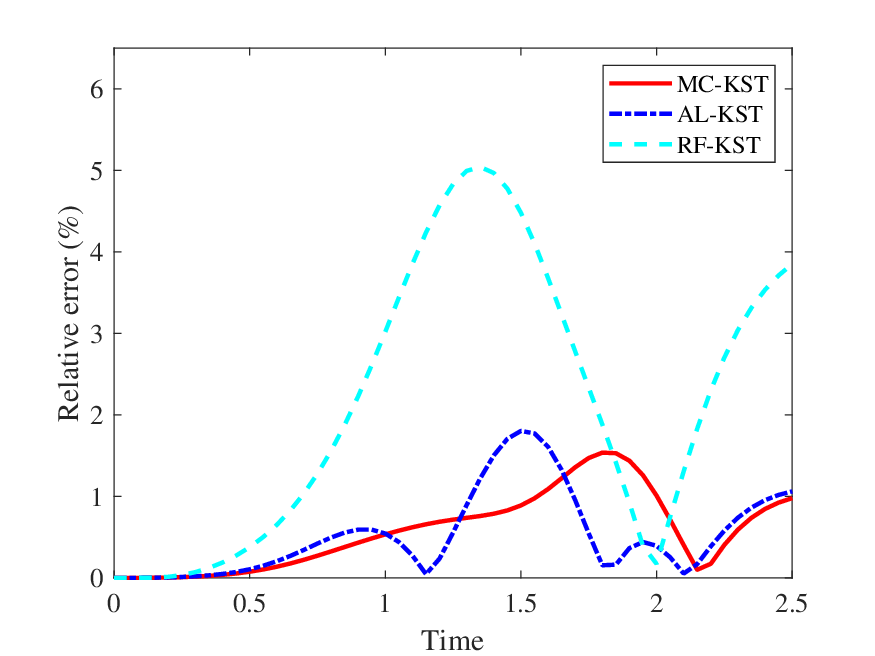}
	\end{minipage}
	\label{fig:ex3_network3_RE_reliability}
}%
\caption{Network reliability results and related relative errors by different methods for edge failure network 3}
\label{fig:example3_network3_result}
\end{figure}

\begin{figure}[!htb]
\centering
\subfigure[Network reliability results]{
	\begin{minipage}{8.0cm}
		\centering
		\includegraphics[scale=0.6, trim=15 0 0 0,clip]{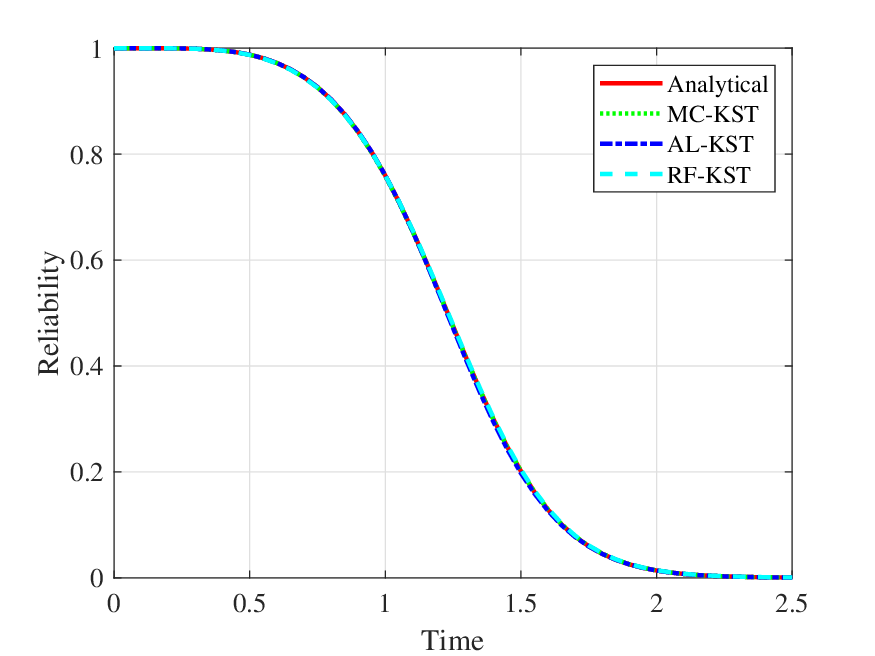}
	\end{minipage}
	\label{fig:ex3_network4_reliability}
}%
\subfigure[Relative errors of network reliability]{
	\begin{minipage}{8.0cm}
		\centering
		\includegraphics[scale=0.6, trim=15 0 0 0,clip]{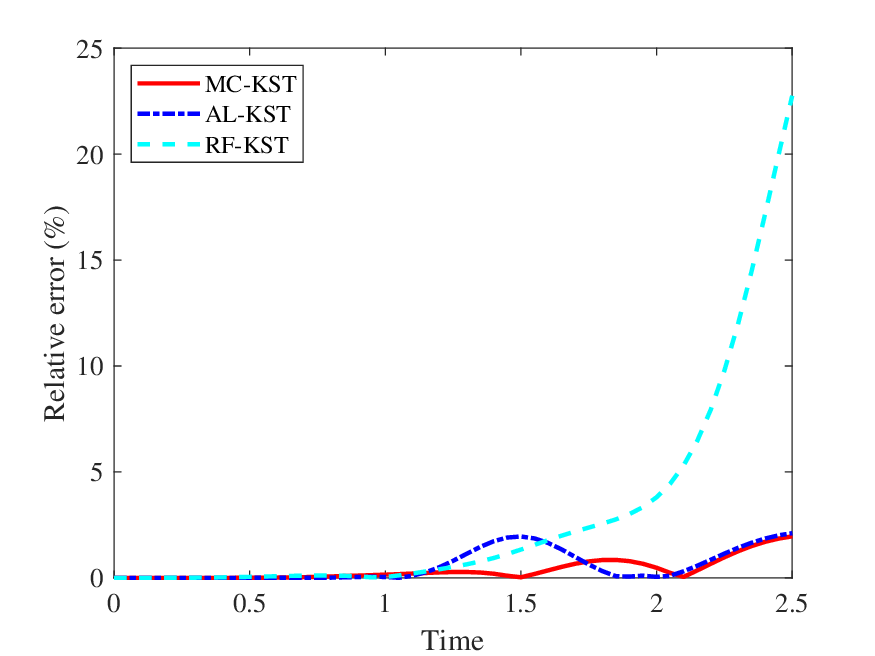}
	\end{minipage}
	\label{fig:ex3_network4_RE_reliability}
}%
\caption{Network reliability results and related relative errors by different methods for edge failure network 4}
\label{fig:example3_network4_result}
\end{figure}

The reliability results and associated relative errors for these four networks obtained by the proposed two methods and \textcolor{black}{RF-KST} are illustrated in Figures \ref{fig:example3_network1_result} to \ref{fig:example3_network4_result}, respectively.
It is evident that all the reliability results predicted by the two proposed methods are in good agreement with the analytical results given by enumeration method.
On the other hand, the results from \textcolor{black}{RF-KST} deviate slightly from the analytical results, particularly noticeable for network 3. 
Moreover, Table \ref{tab:MaxRelativeError_fourNetwork1} presents the maximum relative errors and computational costs for each method.
Among the four networks, the maximum relative error of the two proposed methods occurs in network 4, reaching a maximum value of 2.1106\%, potentially attributed to the larger number of components present.
Nevertheless, the results by the two proposed methods are still in acceptable accuracy.
Additionally, the \textcolor{black}{AL-KST} incurs the maximum computational cost in network 2, primarily due to its dense network structure.
However, when compared to the \textcolor{black}{MC-KST}, the \textcolor{black}{AL-KST} demonstrates over a tenfold improvement in \textcolor{black}{total number of samples} while maintaining similar accuracy levels.
In contrast, \textcolor{black}{RF-KST} exhibits larger maximum relative errors across all networks, peaking at 22.7523\% in network 4. 
These observations underscore the efficacy of the two proposed methods in estimating network reliability for both node and edge failure scenarios with acceptable accuracy, while the proposed \textcolor{black}{AL-KST} further improves the computational efficiency.

\begin{table}[htb!]
	\caption{Maximum relative errors and computational costs by different methods for four networks}
	\setlength\tabcolsep{3pt}
	\label{tab:MaxRelativeError_fourNetwork1}
	\centering 	
		\begin{tabular}{lllllllll}
			\toprule 
			\multirow{2}{*}{Network name}  & \multicolumn{2}{c}{Enumeration} & \multicolumn{2}{c}{\textcolor{black}{MC-KST}}  & \multicolumn{2}{c}{\textcolor{black}{AL-KST}} & \multicolumn{2}{c}{\textcolor{black}{RF-KST}}  \\ \cline{2-9}
			& $RE_{\max}$ & $N$ & $RE_{\max}$ & $N$  & $RE_{\max}$ & $N$ & $RE_{\max}$ & $N$ \\
			\midrule       
			Network 1     & - & $2^{18}$   & 0.3441\% & $50000$ & 0.4024\% & $840$ & 4.8609\% & $840$ \\
			Network 2     & - & $2^{28}$   & 0.5174\%  & $50000$ & 1.0674\% & $4668$ & 5.1115\% & $4668$ \\
			Network 3     & - & $2^{24}$   & 1.5365\% & $50000$ & 1.8030\% & $1968$ & 5.0423\% & $1968$ \\
			Network 4     & - & $2^{30}$   & 1.9597\% & $50000$ & 2.1106\% & $3580$ & 22.7523\% & $3580$ \\
			\bottomrule       
		\end{tabular}
\end{table}

\subsection{\textcolor{black}{Electricity transmission network concerning node failure}}
\begin{figure}[!htb]
	\centering
	\includegraphics[scale=0.85,trim=50 635 80 45,clip] {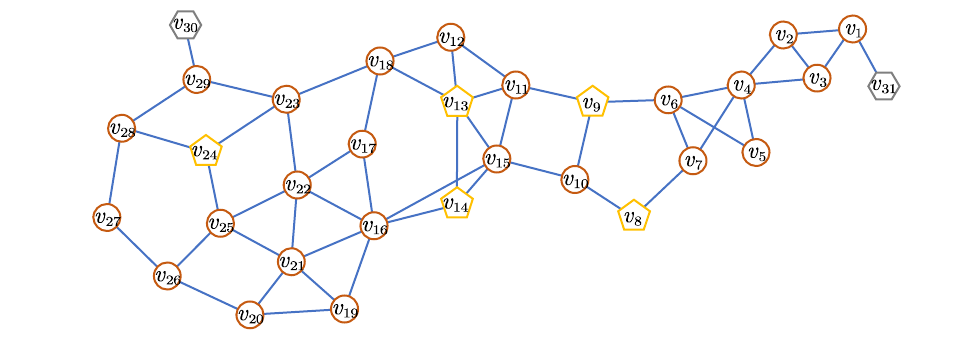}
	\caption{Topology of the Great Britain electricity transmission network}
	\label{fig:GBelectricity_transmission}
\end{figure}

\textcolor{black}{To testify the practical applicability of both proposed methods for node failure scenario, this example considers a reduced representative model of Great Britain electricity transmission network adapted from Ref. \cite{pstca}, as is shown in Figure \ref{fig:GBelectricity_transmission}.
This network concerns 29 nodes prone to failure, which are split to two component classes based on the bus type.
The components denoted by circle markers belong to class 1, and components represented by pentagonal markers belong to class 2.
Lifetimes of class-1 components are assumed to follow an exponential distribution with $\lambda=1$, and lifetimes of class-2 components follow a Weibull distribution with $\alpha_{\mathrm{wbl}}=1$ and $\beta_{\mathrm{wbl}}=2$.
Here, the two-terminal network connectivity between nodes $v_{30}$ and $v_{31}$ is concerned to determine the structure function of this network.
}

\begin{figure}[!htb]
	\centering
	\subfigure[Network reliability results]{
		\begin{minipage}{8.0cm}
			\centering
			\includegraphics[scale=0.6, trim=15 0 0 0,clip]{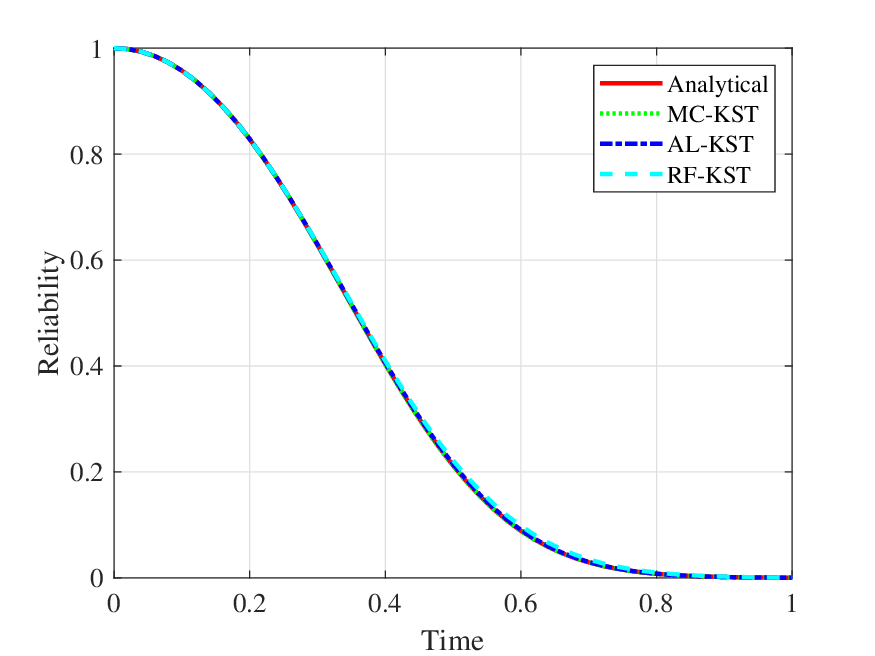}
		\end{minipage}
		\label{fig:ex7_reliability}
	}%
	\subfigure[Relative errors of network reliability]{
		\begin{minipage}{8.0cm}
			\centering
			\includegraphics[scale=0.6, trim=15 0 0 0,clip]{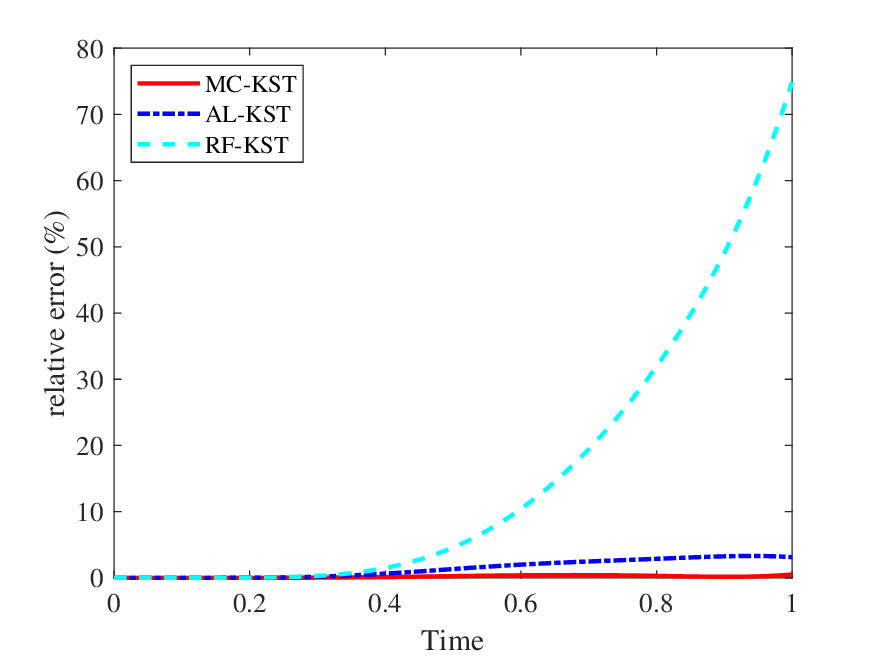}
		\end{minipage}
		\label{fig:ex7_RE_reliability}
	}%
	\caption{Network reliability results and related relative errors by different methods for Great Britain electricity transmission network}
	\label{fig:example7_result}
\end{figure}

\textcolor{black}{Figure \ref{fig:example7_result} presents the network reliability results and associated relative errors for this example estimated by the proposed two methods and RF-KST, alongside the analytical result. 
It can be observed that network reliability results from both proposed methods accord well with the analytical result, while result from the RF-KST performs slightly worse.
The maximum relative errors of MC-KST and AL-KST are 0.5011\% and 3.3127\%, respectively, while that of RF-KST reaches 74.8530\%, showing high accuracy of both proposed methods in this example.
In addition, the proposed MC-KST employs 50000 component lifetime samples to estimate the network reliability, while the AL-KST only adopts 872 samples to predict the reliability with similar accuracy level.}

\subsection{\textcolor{black}{Taiwan academic network concerning edge failure}}
\begin{figure}[!htb]
	\centering
	\includegraphics[scale=0.8,trim=60 205 90 283,clip] {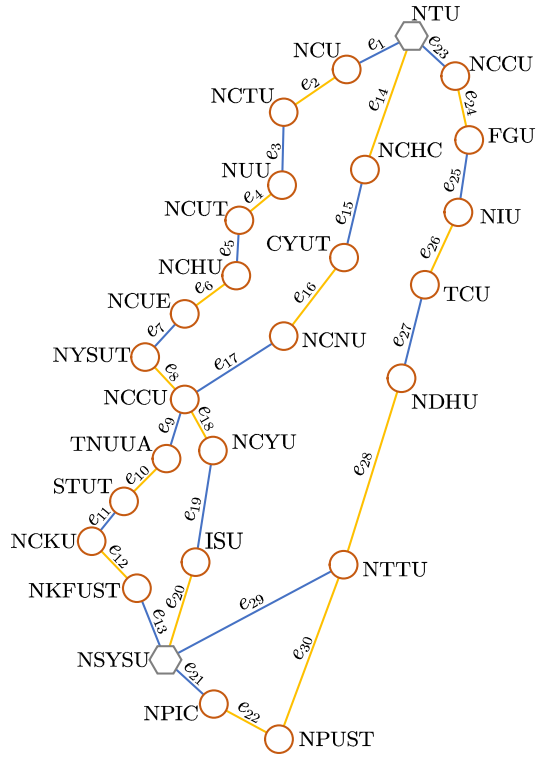}
	\caption{Topology of the Taiwan academic network}
	\label{fig:TANET}
\end{figure}

\textcolor{black}{The last example employs the Taiwan academic network with 30 edges taken from Ref. \cite{lin2012quantifying} to demonstrate the utility of two proposed methods on the practical edge failure scenario.
This network is the main network that links all educational and academic institutions in Taiwan.
The topology of this network is illustrated in Figure \ref{fig:TANET}, where NTU and NSYSU are considered as the source and target nodes, respectively.
Two classes of edge components are concerned here, where the blue lines constitute the class-1 components and yellow lines belong to class-2 components.
We assume the lifetimes of class-1 components follow an exponential distribution with $\lambda=1.2$, and the lifetimes of class-2 components follow a weibull distribution with $\alpha_{\mathrm{wbl}}=1.5$ and $\beta_{\mathrm{wbl}}=3.6$. 
}

\begin{figure}[!htb]
	\centering
	\subfigure[Network reliability results]{
		\begin{minipage}{8.0cm}
			\centering
			\includegraphics[scale=0.6, trim=15 0 0 0,clip]{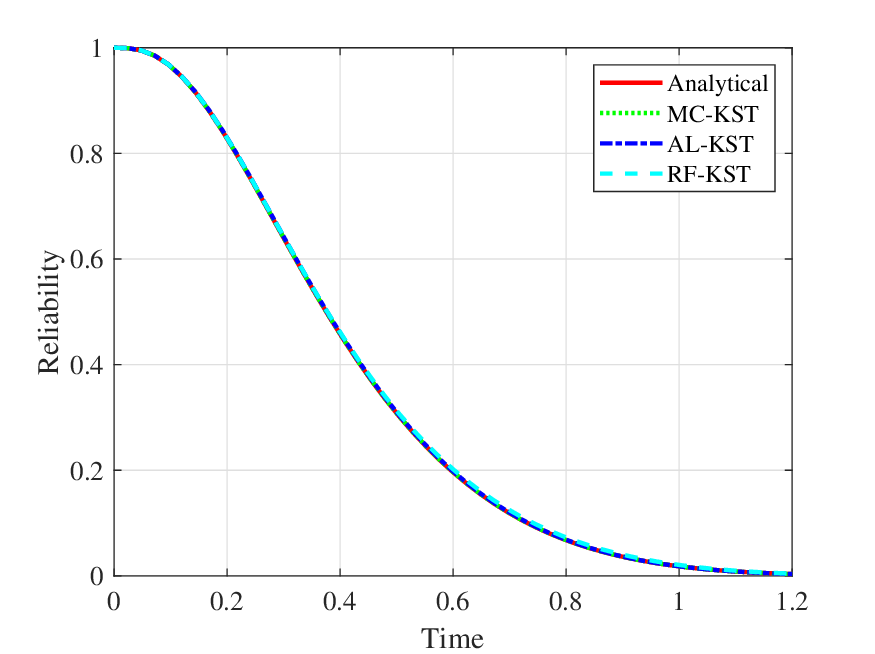}
		\end{minipage}
		\label{fig:ex8_reliability}
	}%
	\subfigure[Relative errors of network reliability]{
		\begin{minipage}{8.0cm}
			\centering
			\includegraphics[scale=0.6, trim=15 0 0 0,clip]{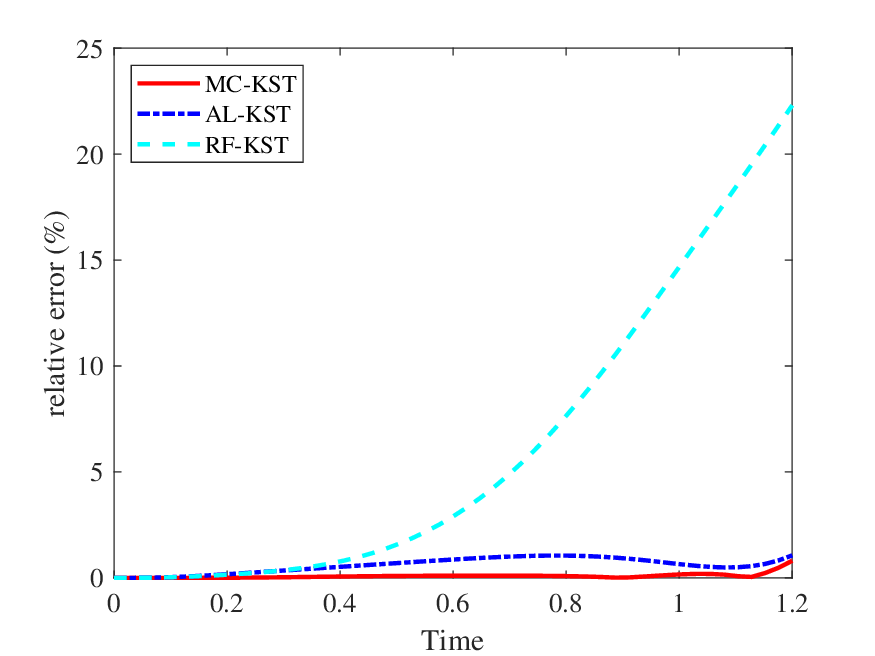}
		\end{minipage}
		\label{fig:ex8_RE_reliability}
	}%
	\caption{Network reliability results and related relative errors by different methods for Taiwan academic network}
	\label{fig:example8_result}
\end{figure}

\textcolor{black}{Results for this example obtained by two proposed methods and RF-KST are given in Figure \ref{fig:example8_result}, together with the analytical result. 
It is evident that compared with the analytical result, both proposed methods enable to provide highly accurate network reliability results, with maximum reliability errors reach 0.8060\% for MC-KST and 1.0607\% for AL-KST.
Although reliability by the RF-KST accords well with the analytical result, the maximum relative error by RF-KST approaches 22.3012\%.
This observation demonstrate the accuracy of both proposed methods for this practical edge failure example.
Furthermore, the AL-KST adopts 756 samples to obtain the reliability result with similar accuracy level of the proposed MC-KST that employs 50000 component lifetime samples, showing improved computational efficiency of the AL-KST.}

\section{\textcolor{black}{Discussion and }concluding remarks}\label{section:conclusion}
In this paper, \textcolor{black}{two novel methods, MC-KST and AL-KST,} are developed to efficiently evaluate network reliability under node failure and edge failure scenarios. 
The proposed \textcolor{black}{MC-KST} offers a unique approach by sampling component lifetimes and employing the K-terminal spanning tree algorithm to accelerate the process of network structure function computation.
Through this process, survival and failure samples of each survival signature combination are collected to approximate the survival signature and assess the network reliability. Additionally, a transformation technique extends the applicability of the \textcolor{black}{MC-KST} to both node failure and edge failure scenarios, enhancing its versatility.
To further enhance the computational efficiency and adaptability across different network topology structures, \textcolor{black}{AL-KST} is proposed by extending \textcolor{black}{MC-KST}.
\textcolor{black}{AL-KST} constructs an RF classifier capable of integrating the network behaviors associated with various network topologies, thus enabling the prediction of structure function values without additional computational burden.
Moreover, the \textcolor{black}{active} learning strategy \textcolor{black}{adopted in AL-KST} allows for the iterative refinement of the RF classifier by sequentially incorporating component lifetime samples, significantly enhancing the prediction accuracy of the RF classifier.

Through the investigation of six diverse network examples \textcolor{black}{and two real-world network cases}, encompassing both node failure and edge failure scenarios, the efficacy of two proposed methods is substantiated.
Results indicate that the proposed methods yield estimates of network reliability with acceptable accuracy. 
Additionally, \textcolor{black}{AL-KST} demonstrates its capacity to further enhance computational efficiency, especially when compared with the performance of the RF classifier without learning.
Notably, the RF classifier trained \textcolor{black}{by AL-KST} exhibits the capability to make accurate predictions regarding the reliability of various variant networks, a capability not achievable with \textcolor{black}{MC-KST and existing MCS-based methods}.

\textcolor{black}{It is worth mentioning that both proposed methods in this paper can be further applied to problems involving both node and edge failures, as well as K-terminal reliability estimation. 
This is because the transformation technique adopted by both methods can be extended to handle scenarios where both nodes and edges are subject to failure. 
In such cases, the transformed edge lifetime is determined as the shortest lifetime among the original edge and its two connected nodes. 
In addition, the K-terminal spanning tree algorithm used in both methods can also determine the K-terminal network connectivity.
We will address these issues in future works.}
	
\textcolor{black}{However, these two methods face limitations in providing accurate reliability analysis for large-scale networks. There are two main reasons for this. First, in larger networks, a limited number of lifetime samples (e.g., 50000) leads to insufficient coverage of all survival signature combinations, affecting the accuracy of network reliability estimation. 
While increasing the sample size (e.g., up to 1 million) could mitigate this issue, it also increases computational time and storage requirements.
Second, the RF classifier struggles with data imbalance due to the constrained size of the training dataset and with predicting in high-dimensional networks where all variables are equally important. 
Therefore, future research will focus on developing MCMC or adaptive MCS methods based on survival signatures to better handle reliability analysis for large-scale networks.
}

\section*{Acknowledgement}
Pengfei Wei and Michael Beer would like to appreciate the support of National Natural Science Foundation of China under grant 72171194.
Chen Ding acknowledges the support of the European Union’s Horizon 2020 research and innovation programme under Marie Sklodowska-Curie project GREYDIENT – Grant Agreement n°955393.
	

	
	\bibliographystyle{elsarticle-num} 
	\bibliography{reference}
	
	
	
	
	
\end{document}